\documentclass{article}

\pdfoutput=1

\usepackage[nonatbib, preprint]{neurips_2025}

\usepackage[utf8]{inputenc} % allow utf-8 input
\usepackage[T1]{fontenc}    % use 8-bit T1 fonts
\usepackage{hyperref}       % hyperlinks
\usepackage{url}            % simple URL typesetting
\usepackage{booktabs}       % professional-quality tables
\usepackage{amsfonts}       % blackboard math symbols
\usepackage{nicefrac}       % compact symbols for 1/2, etc.
\usepackage{microtype}      % microtypography
\usepackage{xcolor}         % colors
\usepackage{booktabs}
\usepackage{multirow}
\usepackage{graphicx}
\usepackage{makecell}
\usepackage{enumitem}
\usepackage{colortbl}
\usepackage{amsmath}
\usepackage{arydshln}
\usepackage{nccmath}
\usepackage{mathtools}

\title{VerIPO: Cultivating Long Reasoning in Video-LLMs via Verifier-Gudied Iterative Policy Optimization}

\author{%
  Yunxin Li$^{1 \ddagger}$\thanks{Equal Contribution, $^\dagger$ Corresponding Author, $^\ddagger$ Project Leader}\hspace{0.5pt}, \hspace{0.5pt} Xinyu Chen$^{1*}$, Zitao Li$^{1}$, Zhenyu Liu$^{1}$, Longyue Wang$^{2}$, Wenhan Luo$^{3}$\\ 
  \textbf{Baotian Hu$^{1 \dagger}$, Min Zhang$^{1}$}\\
  \textsuperscript{1}Harbin Institute of Technology, Shenzhen, China \\
  \textsuperscript{2}Alibaba International Group,
  \textsuperscript{3}Division of AMC and Department of ECE, HKUST\\
  \texttt{liyunxin@stu.hit.edu.cn, \{hubaotian, zhangmin2021\}@hit.edu.cn}\\
  Project Link: \url{https://github.com/HITsz-TMG/VerIPO}
}

\begin{document}

\maketitle

\begin{abstract}
    Applying Reinforcement Learning (RL) to Video Large Language Models (Video-LLMs) shows significant promise for complex video reasoning. However, popular Reinforcement Fine-Tuning (RFT) methods, such as outcome-based Group Relative Policy Optimization (GRPO), are limited by data preparation bottlenecks (e.g., noise or high cost) and exhibit unstable improvements in the quality of long chain-of-thoughts (CoTs) and downstream performance.
    To address these limitations, we propose \textbf{VerIPO}, a Verifier-guided Iterative Policy Optimization method designed to gradually improve video LLMs' capacity for generating deep, long-term reasoning chains.  The core component is \textit{Rollout-Aware Verifier}, positioned between the GRPO and Direct Preference Optimization (DPO) training phases to form the GRPO-Verifier-DPO training loop. This verifier leverages small LLMs as a judge to assess the reasoning logic of rollouts, enabling the construction of high-quality contrastive data, including reflective and contextually consistent CoTs. These curated preference samples drive the efficient DPO stage (7x faster than GRPO), leading to marked improvements in reasoning chain quality, especially in terms of length and contextual consistency. This training loop benefits from GRPO's expansive search and DPO's targeted optimization. Experimental results demonstrate: 1) Significantly faster and more effective optimization compared to standard GRPO variants, yielding superior performance; 2) Our trained models exceed the direct inference of large-scale instruction-tuned Video-LLMs, producing long and contextually consistent CoTs on diverse video reasoning tasks; and 3) Our model with one iteration outperforms powerful LMMs (e.g., Kimi-VL) and long reasoning models (e.g., Video-R1), highlighting its effectiveness and stability.

\end{abstract}

\begin{figure}[htbp]
    \centering
    \includegraphics[width=1.0\linewidth]{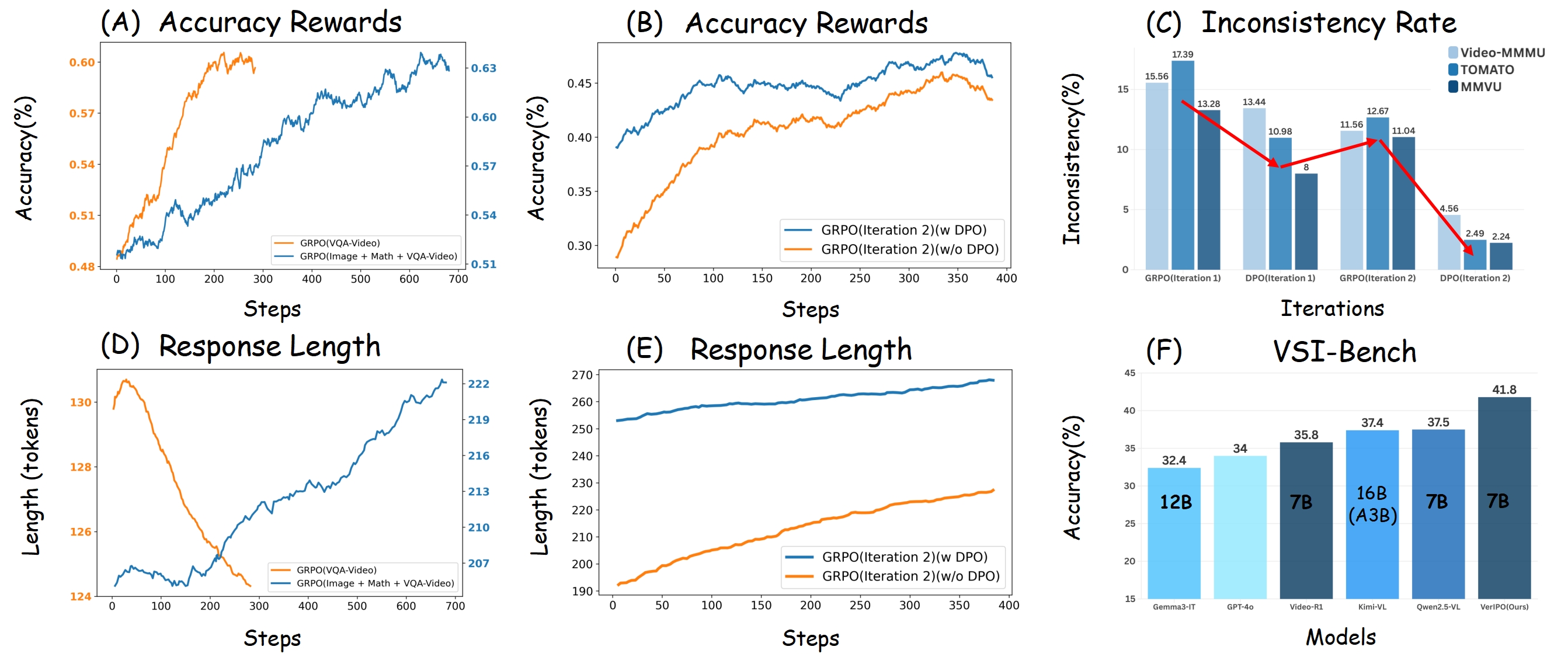}
    \caption{Figures (A, D): Initial GRPO training with different data types shows only utilizing Video-QA data decreases response length. Figures (B, E): Continual GRPO training with/without Verifier-guided DPO (VerIPO) demonstrates VerIPO improves accuracy and response length. Figure (C): Inconsistency rate (thinking vs. final answer) at different stages reveals our method lowers contextual inconsistency of long CoTs while GRPO increases it. Figure (F): Performance on challenging video reasoning dataset VSI-Bench \cite{yang2024think} shows VerIPO (trained with Qwen2.5-VL-7B) outperforms strong LMMs including GPT-4o \cite{hurst2024gpt}, Video-R1 \cite{feng2025videor1reinforcingvideoreasoning}, and Kimi-VL \cite{kimiteam2025kimivltechnicalreport}.}
    \label{fig:absfigure}
    \vspace{-12pt}
\end{figure}

\section{Introduction}

Complex reasoning problems across various domains are often effectively tackled by large models via generating long Chain-of-Thoughts (CoTs) \cite{wei2023chainofthoughtpromptingelicitsreasoning, zhang2024multimodalchainofthoughtreasoninglanguage, zelikman2022starbootstrappingreasoningreasoning, deepseekai2025deepseekr1incentivizingreasoningcapability,li2025perception}, which has demonstrated considerable success in multimodal settings, particularly for challenging tasks like visual math and complex image-text reasoning \cite{wang2025unifiedmultimodalchainofthoughtreward, dong2025insightvexploringlongchainvisual, kimiteam2025kimivltechnicalreport, wu2025stthinkmultimodallargelanguage, xu2025redstardoesscalinglongcot, xiang2024atomthinkslowthinkingframework}.  The activation of this long-form reasoning capability in Large Multimodal Models (LMMs) has been primarily driven by Reinforcement Fine-Tuning (RFT), which combines Supervised Fine-Tuning (SFT) with Long-CoTs data and the application of online reinforcement learning algorithms \cite{tan2025reasonrftreinforcementfinetuningvisual,schulman2017proximalpolicyoptimizationalgorithms, rafailov2024directpreferenceoptimizationlanguage,  yu2025dapoopensourcellmreinforcement, gupta2025ampoactivemultipreferenceoptimization, wu2024selfplaypreferenceoptimizationlanguage, tang2025gametheoreticregularizedselfplayalignment,qwen3}, such as the Group Relative Policy Optimization (GRPO) \cite{shao2024deepseekmathpushinglimitsmathematical} method. Inspired by the success of DeepSeek-R1 \cite{deepseekai2025deepseekr1incentivizingreasoningcapability}, Skywork R1V \cite{chris2025skyworkr1v2multimodalhybrid}, and Vision-R1 \cite{huang2025visionr1incentivizingreasoningcapability}, researchers~\cite{li2025videochatr1enhancingspatiotemporalperception,feng2025videor1reinforcingvideoreasoning, zhang2025tinyllavavideor1smallerlmmsvideo} are actively investigating effective solutions to enhance the long reasoning capabilities of Video Large Language Models (Video-LLMs), aiming to train them to produce effective, long-form reasoning chains for complex video reasoning tasks.

% However, activating the long-form reasoning capability of Video-LLMs faces new challenges:
% \textit{Q1}: Relying on pre-constructed long-CoT video datasets for cold-starting (e.g., Video-R1~\cite{feng2025videor1reinforcingvideoreasoning}) is hindered by the high cost of manual annotation and noise from automatic methods.
% \textit{Q2}: Direct reinforcement fine-tuning (RFT), while improving overall performance compared to supervised fine-tuning~\cite{zhang2024openrftadaptingreasoningfoundation, li2025videochat}, often yields short, shallow reasoning chains lacking deep logical steps, consistent with prior findings\cite{zhang2025tinyllavavideor1smallerlmmsvideo}.
% \textit{Q3}: Direct RFT is prone to producing contextually inconsistent reasoning chains (e.g., correct answers based on wrong thinking), as shown in Figure xxx.
% \textit{Q4}: RFT stages do not consistently improve the accuracy or length of CoTs, particularly with increased temporal frame inputs, and online training does not ensure stable response length increase\cite{zhang2025tinyllavavideor1smallerlmmsvideo}.
% \textit{Q5}: Models using these RFT-based thinking processes (e.g., Video R1 variants) do not achieve consistent performance gains across extensive video reasoning tasks and can sometimes be outperformed by direct inference from instruction-tuned models.
% Overall, substantially improving the comprehensive and deep reasoning capability of Video-LLMs remains an open challenge.

However, activating the long-form reasoning capability of Video-LLMs faces two challenges:
\begin{itemize}[leftmargin=1em]
    \item \textit{Data Preparation Bottleneck}: Employing Long-CoTs video datasets for cold starting (e.g., Video-R1~\cite{feng2025videor1reinforcingvideoreasoning}) is hindered by the high cost of manual annotation and noise from automatic methods.
    
    \item \textit{Limitations and Instability of Online RL}:
    First, while improving overall performance compared to SFT~\cite{li2025videochatr1enhancingspatiotemporalperception}, direct RL training on Video-LLMs often yields short, shallow reasoning chains that lack deep logical steps, consistent with prior findings \cite{zhang2025tinyllavavideor1smallerlmmsvideo}.
    Second, GRPO is prone to making models that often produce contextually inconsistent reasoning chains (Figure (C)), e.g., ``correct answers based on wrong thinking''.
    Third, RL training does not consistently improve the accuracy or length of CoTs, particularly with increased temporal frame inputs, and online training does not ensure a stable increase in response length \cite{zhang2025tinyllavavideor1smallerlmmsvideo} (Figure (D)).
    Finally, models with long reasoning processes (e.g., Video-R1) show inconsistent performance gains across video reasoning tasks, occasionally performing worse than direct inference by instruction-tuned models.
\end{itemize}
\vspace{-4pt}

Hence, substantially improving the deep reasoning capability of Video-LLMs still remains an open challenge.
To address these limitations, we propose \textbf{VerIPO}, an online rollout-aware $\textbf{Ver}$ifier-guided $\textbf{I}$terative $\textbf{P}$olicy $\textbf{O}$ptimization algorithm. Our approach forgoes large Long-CoT SFT datasets for cold starting, instead directly applying RL to cultivate long reasoning capability in Video-LLMs gradually. Specifically, the \textit{rollout-aware Verifier}, a core component of {VerIPO}, is strategically positioned between the GRPO and DPO training phases to form the GRPO-Verifier-DPO loop.
This verifier leverages small LLMs to critically assess the reasoning quality and contextual consistency of generated CoTs. Through this quality assessment, the verifier intelligently selects high-quality contrastive samples, facilitates the construction of reflective and contextually consistent CoTs, particularly based on GRPO rollouts. These contrastive samples are then utilized to train the model via an efficient DPO stage, which was empirically found to be 7x faster than GRPO (as shown in \ref{training_speed}) and support the targeted optimization of the reasoning path compared to GRPO. The verifier can also progressively prune simple training examples that the model has mastered, thus contributing to speeding up the overall training process. During training, we incorporate diverse video QA datasets complemented by high-quality image and textual math datasets used particularly in the initial phase, to enhance the range of logical paths explored by models.

% During training, we incorporate a diverse set of video QA datasets, complemented by high-quality image and textual math datasets at the beginning to enhance the diversity of potential logical paths explored by the language model. and identifies simple training instances that the model has mastered

We conduct extensive experiments on four video reasoning and two long video understanding benchmarks, e.g., VSI-Bench~\cite{yang2024think} and Video-MME~\cite{fu2024videomme}. Our experimental results show that \textbf{VerIPO} achieves consistent and significant performance improvements and outperforms larger Video-LLMs and powerful RFT models Video-R1  and Kimi-VL-Thinking \cite{kimiteam2025kimivltechnicalreport}. It highlights the effectiveness and stableness of VerIPO in cultivating the long-reasoning ability of Video-LLMs. Compared to RFT with long-CoTs dataset as a cold start, our approach consistently generates longer responses and improves the quality of generated long CoTs, e.g., contextual consistency and low repetition.
Our contributions can be summarized as follows:

\begin{itemize}[leftmargin=1em]

 \item We propose \textbf{VerIPO}, a novel Verifier-guided Iterative Policy Optimization algorithm designed to improve the long reasoning capability of Video-LLMs. The method enhances rollout data utilization via the embedded Verifier system and efficient DPO, enabling the model to realize improvement via effective learning from its online running experience.
 
 \item The rollout-aware Verifier analyzes and refines generated rollout data into high-quality, reflective contrastive samples, which are essential for continuously improving the model's long-term reasoning capabilities during the DPO training stage.
 
% \item Experimental results demonstrate that \textbf{VerIPO} effectively improves long reasoning performance on general video understanding tasks. Our trained models consistently generate longer and more accurate reasoning chains, surpassing previous RL-trained thinking models (e.g., Video-R1, Kimi-VL-Thinking), original Video-LLMs (excedding direct-answer Qwen2.5-VL-7B-Instruct on four benchmarks), larger models (latest LMMs with $>$ 11B).

\item Experimental results demonstrate that VerIPO significantly improves long reasoning performance on \textit{general} video QA tasks. Our trained models consistently generate long and accurate reasoning chains, outperforming RL-trained thinking models (Video-R1 and Kimi-VL-Thinking), direct-answer models (like Qwen2.5-VL-7B \cite{bai2025qwen25vltechnicalreport} on four benchmarks), and larger latest LMMs ($>$ 11B).

\end{itemize}

\section{Related Work}

\paragraph{Large Multimodal Models for Video Reasoning}
Video reasoning is the core capability of Large Multimodal Models (LMMs), enabling understanding of interactions, dependencies, and inference over dynamic content~\cite{10598361, 10887014, zhang2025tinyllavavideor1smallerlmmsvideo, zheng2025villavideoreasoningsegmentation}. Specifically, spatial reasoning models object relationships and scene layouts within frames, while temporal reasoning captures motion, causality, and sequence across frames~\cite{ouyang2025spatialr1enhancingmllmsvideo, daxberger2025mmspatialexploring3dspatial, ray2025satdynamicspatialaptitude, liu2025spatialcotadvancingspatialreasoning}.
Early Video-LLMs focused on short videos using pre-trained image~\cite{dosovitskiy2021imageworth16x16words, oquab2024dinov2learningrobustvisual, radford2021learningtransferablevisualmodels} or video encoders~\cite{arnab2021vivitvideovisiontransformer, liu2021videoswintransformer, neimark2021videotransformernetwork} with frozen language models~\cite{dai2023instructblipgeneralpurposevisionlanguagemodels, li2022mplugeffectiveefficientvisionlanguage, li2024videochatchatcentricvideounderstanding, maaz2024videochatgptdetailedvideounderstanding, zhang2023videollamainstructiontunedaudiovisuallanguage}. Recent efforts target long-form video understanding with complex temporal and multimodal reasoning~\cite{fei2024videoofthoughtstepbystepvideoreasoning, feng2025videor1reinforcingvideoreasoning, zhang2025tinyllavavideor1smallerlmmsvideo, zheng2025villavideoreasoningsegmentation, liu2025videomindchainofloraagentlong, chen2025visrlintentiondrivenvisualperception, liu2025othinkmr1stimulatingmultimodalgeneralized}.
To handle long contexts, methods adopt hierarchical temporal attention and larger context windows~\cite{liu2025videomindchainofloraagentlong, wei2025videoropemakesgoodvideo}, or compress visual inputs via event-level abstraction~\cite{zhang2025tinyllavavideor1smallerlmmsvideo, chen2024mecdunlockingmultieventcausal}. 
% Advances in spatial reasoning include 3D-aware representations, grounding, and coordinate-based logic~\cite{ouyang2025spatialr1enhancingmllmsvideo, daxberger2025mmspatialexploring3dspatial, ray2025satdynamicspatialaptitude, liu2025spatialcotadvancingspatialreasoning}.
Recent multimodal fusion integrates audio and motion cues for improved understanding in videos~\cite{chen2025visrlintentiondrivenvisualperception, zhao2025r1omniexplainableomnimultimodalemotion, liu2025visualrftvisualreinforcementfinetuning}. Reinforcement learning guides perception and reasoning, aiding in interpretability and intent modeling~\cite{deng2025boostinggeneralizationreasoningvision, liu2025othinkmr1stimulatingmultimodalgeneralized, liu2025segzeroreasoningchainguidedsegmentation}.
Recent work explores structured outputs, intention-driven attention, and stepwise reasoning~\cite{chen2025visrlintentiondrivenvisualperception, yang2025r1onevisionadvancinggeneralizedmultimodal, huang2025visionr1incentivizingreasoningcapability, peng2025lmmr1empowering3blmms} for fine-grained grounding and spatiotemporal segmentation.

% Some unified benchmarks~\cite{meng2025mmeurekaexploringfrontiersmultimodal} aim to standardize evaluation across tasks. 
% 可补充 benchmark 名称和介绍

\paragraph{Reinforcement Learning for Multimodal Reasoning}
Reinforcement learning (RL) has become a pivotal approach for aligning LLMs and LMMs with complex reasoning objectives. Foundational policy optimization algorithms, such as Proximal Policy Optimization (PPO), Direct Preference Optimization (DPO), and Group Relative Policy Optimization (GRPO), have been instrumental in this domain~\cite{schulman2017proximalpolicyoptimizationalgorithms, rafailov2024directpreferenceoptimizationlanguage, shao2024deepseekmathpushinglimitsmathematical}. Further advancements have enhanced training stability and efficiency~\cite{yu2025dapoopensourcellmreinforcement, gupta2025ampoactivemultipreferenceoptimization, wu2024selfplaypreferenceoptimizationlanguage, tang2025gametheoreticregularizedselfplayalignment}.
A critical challenge in popular RL training is the ``cold start" problem, where initializing models without prior guidance can lead to suboptimal performance. To mitigate this, Reinforcement Fine-Tuning (RFT) has been proposed, wherein models undergo preliminary SFT on curated datasets to stabilize subsequent RL training phases~\cite{liu2025visualrftvisualreinforcementfinetuning, zhang2024openrftadaptingreasoningfoundation, tan2025reasonrftreinforcementfinetuningvisual, shi2025efficientreinforcementfinetuningadaptive, chen2025suitabilityreinforcementfinetuningvisual, li2025videochatr1enhancingspatiotemporalperception, wang2025sotalessmctsguidedsample, luo2025guir1generalistr1style, wang2025unifiedmultimodalchainofthoughtreward, xing2025echoinkr1exploringaudiovisualreasoning}. Additionally, some verifiers, designed to assess and guide the quality of generated outputs, have proven beneficial. These verifiers assist in filtering and selecting high-quality training samples, thereby enhancing the efficiency and effectiveness of the training process~\cite{chen2025visrlintentiondrivenvisualperception, zhao2025r1omniexplainableomnimultimodalemotion, sun2025mmverifyenhancingmultimodalreasoning, wang2024piecingtogetherverifyingmultihop}.

\section{Preliminary}

% In this section, we summarize the key formulations behind \textit{Direct Preference Optimization (DPO)} \cite{rafailov2024directpreferenceoptimizationlanguage} and \textit{Group Relative Policy Optimization (GRPO)} \cite{shao2024deepseekmathpushinglimitsmathematical}.

\paragraph{Direct Preference Optimization (DPO)}

DPO \cite{rafailov2024directpreferenceoptimizationlanguage} optimizes a policy $\pi_\theta$ to prefer a response $y_+$ over $y_-$ for a given input $x$, with regularization from a reference model $\pi_{\text{ref}}$. The core loss function is:

\vspace{-5pt} % Adjust the space below the equation
\begin{equation}
\mathcal{L}_{\text{DPO}}(\pi_\theta; \pi_{\text{ref}}) = -\mathbb{E}_{(x, y_+, y_-) \sim \mathcal{D}} \left[ \log \sigma \left( \beta \log \frac{\pi_\theta(y_+ \mid x)}{\pi_{\text{ref}}(y_+ \mid x)} - \beta \log \frac{\pi_\theta(y_- \mid x)}{\pi_{\text{ref}}(y_- \mid x)} \right) \right],
\end{equation}

\vspace{-5pt} % Adjust the space after the equation
where $\sigma(\cdot)$ is the sigmoid function, $\beta > 0$ is a temperature parameter, and $\mathcal{D} = \{(x, y_+, y_-)\}_{i=1}^N$ is a static dataset of comparisons sampled from human preference distribution.
This can be interpreted as minimizing the binary cross-entropy between a pairwise preference label and the log odds induced by the policy relative to the reference.
This approach is a \underline{\textit{targeted and fast}} optimization for models.

\paragraph{Group Relative Policy Optimization (GRPO)}

%\vspace{-5pt} % Adjust the space before the equation
For a given input $q$, the model generates a group of $G$ responses $\{y_1, y_2, \ldots, y_G\}$ sampled from the current policy $\pi_\theta$. Each response $y_i$ is assigned a reward $r(y_i)$, typically derived from human feedback or automated evaluation metrics. Following outcome supervision method, the group mean reward $\mu$ and standard deviation $\sigma$ are computed to obtain the advantage score:
\vspace{-2pt} % Adjust the space below the equation
\begin{equation}
\mu = \frac{1}{G} \sum_{i=1}^G r(y_i), \quad \sigma = \sqrt{\frac{1}{G} \sum_{i=1}^G (r(y_i) - \mu)^2}, \quad \hat{A}_{i,t} = \frac{r(y_i) - \mu}{\sigma}.
\end{equation}

\vspace{-5pt} % Adjust the space before the next equation
With the score computed, GRPO \cite{shao2024deepseekmathpushinglimitsmathematical} updates the policy by maximizing the following objective:
\vspace{-2pt} % Adjust the space below the equation
{
\begin{align}
 &\mathcal{L}_{\text{Advantage}}(\pi_\theta) = \mathbb{E}_{q \sim \mathcal{P}(Q), \{y_i\}_{i=1}^G \sim \pi_{\theta_{\text{ref}}}(y_{i,t} \mid q, y_{i,<t}) }  \nonumber \\ & 
\frac{1}{G} \sum_{i=1}^G \frac{1}{|y_i|} \sum_{t=1}^{|y_i|} \left\{
\min \left[
\frac{\pi_\theta(y_{i,t} \mid q, y_{i,<t})}{\pi_{\theta_{\text{ref}}}(y_{i,t} \mid q, y_{i,<t})} \hat{A}_{i,t}, \;
\text{clip}\left( \frac{\pi_\theta(y_{i,t} \mid q, y_{i,<t})}{\pi_{\theta_{\text{ref}}}(y_{i,t} \mid q, y_{i,<t})}, 1 - \epsilon, 1 + \epsilon \right) \hat{A}_{i,t}
\right]
\right\} \\ 
&\mathcal{L}_{\text{GRPO}}(\pi_\theta) = \mathcal{L}_{\text{Advantage}}(\pi_\theta) - \beta \, \mathrm{D_{KL}}\left[ \pi_\theta \| \pi_{\text{ref}} \right]
\end{align}
}

\vspace{-5pt} 
where $\epsilon$ is a hyperparameter controlling the clipping range, $\beta$ is the temperature parameter, and $\pi_{\theta_{\text{ref}}}$ is the policy before the update. This approach allows the model to focus on generating responses that are relatively better within a group, promoting \underline{\textit{wide yet slow}} exploration in the generation space. 

%However, in our work, we use $\mathcal{L}_{\text{GRPO}}^{sim}(\pi_\theta)$ only.

\section{VerIPO: Verifier-Guided Iterative Policy Optimization}

\subsection{Overview}

We introduce VerIPO, an iterative policy optimization approach specifically designed to enhance the long reasoning capability of Video-LLMs. The method follows an iterative process:
1) \textit{Initial Policy Exploration:} We first apply GRPO to the instruction-tuned Qwen2.5-VL, utilizing diverse accuracy rewards tailored for various video task output formats. 2) \textit{Sample Curation with Verifier:} A Verifier component analyzes the GRPO rollouts to produce high-quality, long reasoning paths that lead to accurate answers as positive (chosen) samples. It also selects challenging, incorrect reasoning paths as hard negative (rejected) samples. 3) \textit{Policy Refinement with DPO}: These curated contrastive samples are then used to fine-tune the model via DPO. The DPO efficiently refines the model's policy, encouraging the generation of better reasoning paths in a controllable direction. 

%This DPO-based training provides process supervision on the reasoning path itself, which is faster compared to rule-based GRPO alone.

% In this paper, we mainly design an iterative policy optimization approach VerIPO, directly used for open-source Qwen2.5-VL to improve its long reasoning capability on video reasoning tasks. We first apply GRPO for the instruction-tuned Qwen2.5-VL \cite{bai2025qwen25vltechnicalreport} with diverse accuracy rewards, which are designed for video tasks with different outputting formats. Then, we introduce a Verifier to analyze the rollouts of GRPO and produce high long-term reasoning paths with accurate answers. These high-quality positive samples and selective hard rejected samples are used to continually train original models with DPO. This process aims to continually activate models to produce better reasoning paths with controllable direction. Compared to rule-based GRPO, the training of DPO is fast and it can achieve process supervision of reasoning path via the contrastive sample pairs. The whole GRPO-Verfier-DPO training loop will be guided by the Verifier to control the quality of training samples.

\subsection{GRPO}

Following the GRPO algorithm from DeepSeek-R1~\cite{deepseekai2025deepseekr1incentivizingreasoningcapability}, we employ two types of rewards: accuracy and format. The accuracy reward $r_a$  is scaled within the range $[0,1]$, while the format reward $r_f$ is bounded within $[0,0.5]$. 
The calculation of accuracy reward $r_a$ depends on the type of question posed in the input prompt. For mathematical questions, we employ Math-Verify\footnote{https://github.com/huggingface/Math-Verify} to parse the answer from the model's output and compare it against the ground truth $GT$, yielding a binary reward (1 for correct, 0 for incorrect). Similarly, for multiple-choice questions, $r_a$ is assigned a value of 1 if the model's selected option aligns with the ground truth $GT$ and 0 otherwise.
As for distance estimation tasks, we utilize the Mean Relative Accuracy (MRA) metric, as proposed in VSI-Bench~\cite{yang2024think}, which provides a continuous reward value between 0 and 1.
The format reward $r_f$ is binary (0.5 for adherence, 0 for non-adherence), contingent upon whether the model's response conforms to the predefined \textit{<think>...</think><answer>...</answer>} structure. The accuracy rewards are presented as
\vspace{-2pt}
\begin{equation}
r_a = \begin{cases}
1 & \text{if } Q_{type} \in \{\text{Math, MC}\} \text{ and Answer matches \textit{GT}} \\
0 & \text{if } Q_{type} \in \{\text{Math, MC}\} \text{ and Answer does not match \textit{GT}} \\
MRA(Output, GT) & \text{if } Q_{type} = \text{Distance Estimation}
\end{cases}
\end{equation}
To broaden the model's exploration capabilities and enhance learning flexibility, we remove the KL divergence during the GRPO training process. 
Furthermore,  we encountered an empirical observation consistent with findings reported in DAPO~\cite{yu2025dapoopensourcellmreinforcement}. 
As training progressed, the number of samples with an accuracy of 1 continually increased. 
These samples have an advantage of 0 and result in no gradient for policy updates, which suppressed the gradient signals during the model's training process.  To mitigate this phenomenon and maintain robust gradient flow, we integrate the online filter strategy\cite{meng2025mmeurekaexploringfrontiersmultimodal} to exclude zero-advantage samples from the training batches.

\begin{figure}[t]
    \centering
    \includegraphics[width=0.95\linewidth]{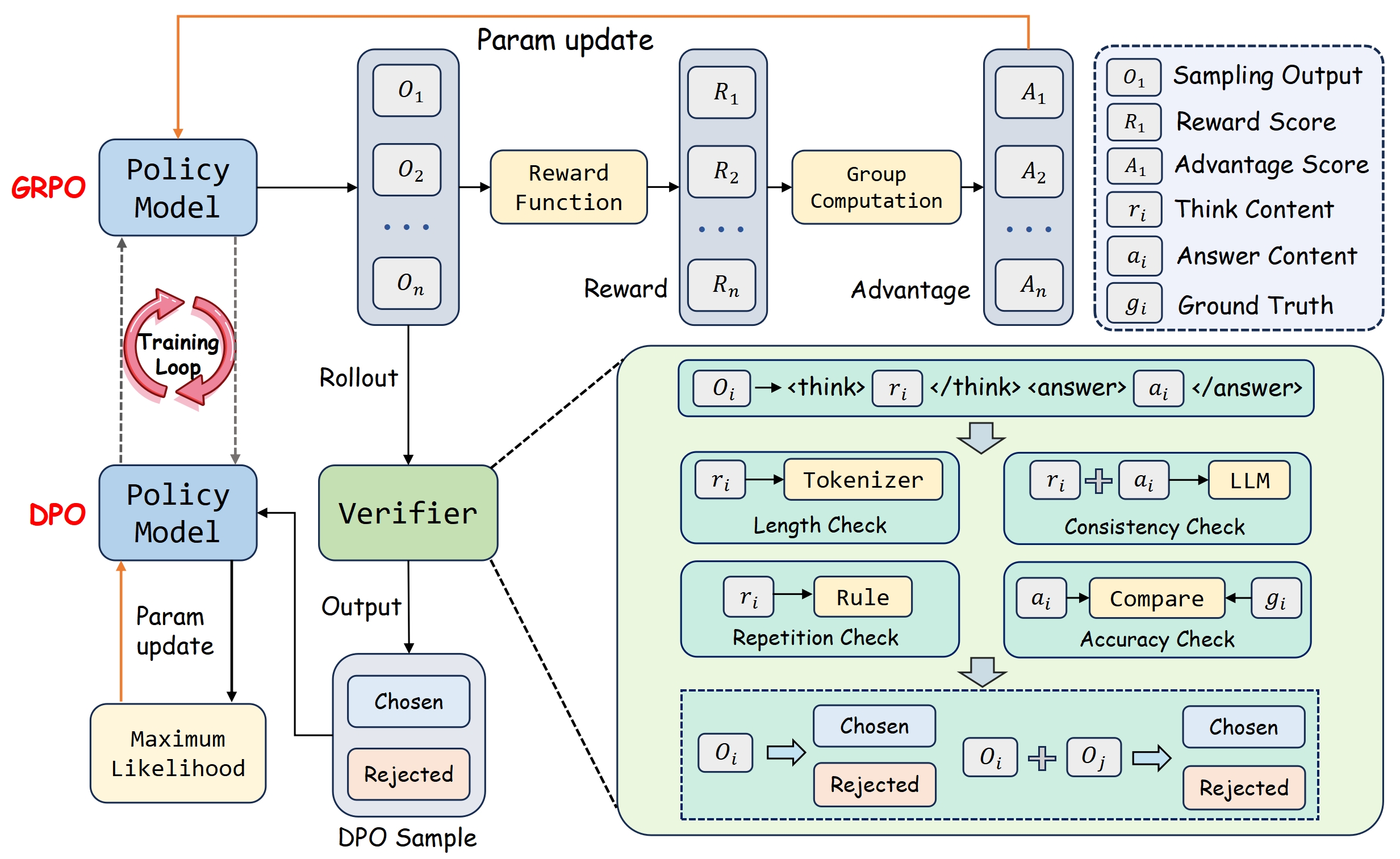}
    \caption{Overview of VerIPO workflow. This training loop is guided by the Verifier's continuous evaluation and selection of training samples. The optimization process progressively improves the model's long reasoning capability by learning from high-quality and informative reasoning examples.}
    \label{fig:model_arc}
    \vspace{-18pt}
\end{figure}

\subsection{Rollout-Aware Verifier}

To address the limitation of outcome-based GRPO in optimizing reasoning paths, we introduce a rollout-aware Verifier that analyzes online rollouts to generate high-quality preference data, continuously guiding the model to generate long-term, high-quality reasoning paths. As shown in Figure~\ref{fig:model_arc}, for a given rollout $o_i$, we employ regular expressions to extract both the thought content $r_i$ and answer $a_i$. The verifier encompasses
four-aspect quality assessment to select high-quality long-CoT samples:

\textbf{Accuracy Check} determines the correctness of the extracted answer content, employing the accuracy function used in GRPO. Notably, for distance estimation tasks, an answer is deemed correct if its MRA surpasses a threshold of 0.6. Samples that successfully pass these verification stages are considered for constructing the positive sample.

\textbf{Consistency Check} evaluates the coherence between the reasoning process and the final answer. It uses the Qwen3-8B to extract the reasoning answer  $a^r_i$ from the response $r_i$, given the original question. A rule-based method then verifies if $a^r_i$ matches the predicted answer $a_i$ to assess reasoning consistency. This checking stage is mainly used to detect the right reasoning with the correct answer and remove the sample with an error reasoning path.

\textbf{Repetition Check} applies rule-based methods to detect any sentence-level repetition within the thought content $r_i$, penalizing responses that exhibit repetitive patterns.

\textbf{Length Check} is applied on the extracted thinking content $r_i$ of rollouts, which assigns higher priority to longer samples for those as positive examples.

%The initial checking stages serve to select rollout samples exhibiting high quality and a coherent, potentially long-term, reasoning process. 

Following this selection, we construct contrastive pairs for DPO training.
Samples are initially categorized based on their average accuracy reward $r_a^{avg} = \frac{1}{N}\sum_{i=1}^{N}a_i$, where $N$ refers to the number of sampling per query. This classification guides the data construction process.
For training samples where the model consistently produces incorrect rollouts ($r_a^{avg}=0$), their high-quality long-term reasoning is generated using Gemini-2.5-Flash. These will help models explore deep reasoning for challenging questions. Conversely, rollouts with perfect accuracy ($r_a^{avg}=1$) are regarded as simple samples and generally excluded from preference pairs during the DPO stage. Then, the contrastive preference dataset is specifically constructed using the following strategies:

\textbf{Single-Turn Preference Pairs:} Negative examples are randomly selected from incorrect rollouts. Positive examples are constituted by the longest rollouts that successfully passed all preceding quality and consistency checks, thereby rewarding thorough correct reasoning processes.

    % \item \textbf{Dual-Correct Preference Pairs (for High-Performing Prompts):} For prompts where the model performs well ($r_a^{avg} \ge 0.75$), we construct pairs where both the chosen and rejected rollouts are verified as correct. Here, preference is based on other quality attributes: shorter correct rollouts serve as rejected samples, while the longest correct rollouts are designated as chosen, encouraging detailed and exhaustive correct answers.

\textbf{Repetition Penalty Pairs:} To specifically address and penalize repetitive outputs, rejected examples are rollouts flagged by the Verifier for containing sentence-level repetitions. To maintain training stability, these negative samples are truncated to a maximum of 1024 tokens. Corresponding chosen examples are the longest verified correct rollouts devoid of such repetitions.

\textbf{Reflective Preference Pairs:} For prompts where the model initially struggled ($r_a^{avg} \le 0.25$), these pairs guide the model in self-correcting its reasoning. Rejected examples concatenate two or more distinct incorrect rollouts, as shown in Figure \ref{fig:model_arc}. Chosen examples combine initial incorrect rollout sequences with a correct one at last, linked by reflective phrases to simulate refined reasoning.

\textbf{Inference Consistency Pairs:} This category enhances the alignment between the model's reasoning ("think content") and the final answer. Rejected examples consist of rollouts where the think content and final answer are incongruent (e.g., incorrect reasoning with a correct answer, or correct reasoning with an incorrect answer). Rollouts with correct reasoning but incorrect answers are rejected against a chosen sample using the correct reasoning path with the extracted reasoning answer ($a^r_i$). Rollouts with incorrect reasoning but correct answers are rejected, with a randomly selected contextually consistent response with the right answer serving as the chosen sample.

% Reasoning-answer inconsistencies inform DPO preference pair construction.Specifically, for cases with incorrect answers but correct reasoning, we construct a preference pair: the rejected sample is the original inconsistent rollout, while the chosen sample uses the correct reasoning path but replaces the original incorrect answer with the extracted reasoning answer ($a^r_i$), thus presenting a consistent version.
% For incorrect reasoning with a correct answer, we randomly select a consistent right generation as the positive sample.
\begin{table}[t]
\renewcommand{\arraystretch}{0.8}
  \caption{Training data and hyperparameters across different stages.}
  \label{tab:training_stage}
  \centering
  \resizebox{1.0\textwidth}{!}{%
  \begin{tabular}{lcccc}
    \toprule
    % \textbf{Stage} & \textbf{\makecell{Reasoning \\ Activation}} & \textbf{\makecell{Video \\ Generalization}} & \textbf{DPO} & \textbf{GRPO2} \\
    \textbf{Stage} & \textbf{Reasoning Activation} & \textbf{Group-Slow-Search} & \textbf{Pair-Fast-Align} & \textbf{Group-Slow-Search} \\
    \midrule
    % \textbf{Framework} & \textbf{OpenRLHF} & \textbf{OpenRLHF} & \textbf{trl} & \textbf{OpenRLHF} \\
    \textbf{Algorithm} & \textbf{GRPO} & \textbf{GRPO} & \textbf{DPO} & \textbf{GRPO} \\
    \midrule
    \textbf{Data} & \makecell{Long Document (1k) \\ Math-Text (30k) \\ Reasoning-Image (39K)} &
    \makecell{Science-Image (4K) \\ Spaital-Image (9k) \\ General-Image (10K) \\ VQA-Video (24k)} & \makecell{Rollouts of \\ VQA-Video}  & \makecell{VQA-Video}\\
    \midrule
    \textbf{Gloabl Batch Size} & 128 & 64 & 32 & 64 \\
    \textbf{Rollout Batch Size} & 64 & 64 & - & 64 \\
    \textbf{Learning Rate} & 1e-6 & 1e-6 & 5e-7 & 5e-7 \\
    \textbf{Rollout Responses per Query} & 8 & 8 & - & 8 \\
    \textbf{Sampling Temperature} & 1.0 & 1.0 & - & 1.0 \\
    \textbf{DPO Beta ($\beta$)} & - & - & 0.1 & - \\
    \bottomrule
  \end{tabular}
  }
\vspace{-10pt}
\end{table}

%Science-Image(4K) \\ Spaital-Image(9k) \\ General-Image(10K) \\ VQA-Video(24k)

% \end{enumerate}
This multi-faceted checking and data construction pipeline yields a rich and diverse preference dataset, specifically engineered to support robust and fast DPO training focused on improving the model's reasoning length, self-reflection, and logical consistency.
\vspace{-2pt}

\subsection{DPO and Training Loop}

Based on the model from the previous GRPO round, DPO training is performed on contrastive data generated by the rollout-aware verifier. The visual encoder is kept frozen throughout this process, and further training parameter configurations are detailed in Table~\ref{tab:training_stage}.

The training loop follows a curriculum learning approach to gradually activate the LMMs' long-term reasoning ability in video. This begins with simple-modality data (text-only or image QA) for initial \textit{reasoning activation} with GRPO, followed by the GRPO training using image and video QA data, as shown in Table~\ref{tab:training_stage}.
Then, the whole GRPO-Verifier-DPO pipeline continuously enhances the model's long-term reasoning capability and gradually stabilizes its performance on video reasoning, iteratively pushing towards the model's inherent reasoning limit. During the iterative process, we will gradually discard 80\% of the simple examples ($r_a^{avg}=1$) from the previous GRPO training process to reduce the training time of models. The entire training process equips LMMs with robust long-chain reasoning ability with slow-search GRPO and fast-align DPO.

\vspace{-2pt}

\section{Experiment}

\subsection{Experiment Setup}

\textbf{Baseline}.
We compare VerIPO against various SFT and RL baselines. Direct-answer models (SFT, size $>7B$) respond without an explicit reasoning process, while reasoning-answer models generate a reasoning process before answering. Direct-answer baselines include SOTA models like LLaMA-3.2-V \cite{meta2024llama32}, Gemma-3-IT \cite{gemmateam2025gemma3technicalreport}, Kimi-VL-A3B \cite{kimiteam2025kimivltechnicalreport}, Qwen2.5-VL-Instruct \cite{bai2025qwen25vltechnicalreport}, and others. Reasoning-answer baselines include Kimi-VL-A3B-Thinking \cite{kimiteam2025kimivltechnicalreport} and Video-R1 \cite{feng2025videor1reinforcingvideoreasoning}.

\textbf{Training Details}.
Our GRPO algorithm is implemented using the OpenRLHF framework, and DPO training uses the TRL framework with a $\beta$ value of 0.1. Based on Qwen2.5-VL-7B, we conduct experiments on eight NVIDIA A800-80G GPUs with a maximum of 64 frames and 128*28*28 resolution. The global training batch size is set to 64, with a rollout training batch size of 64 and 8 rollout responses per query, the sampling temperature is fixed at 1.0, and the maximum output length is 4096 tokens. The learning rate is set to 1e-6. Detailed settings are shown in Table \ref{tab:training_stage} and \textcolor{blue}{\ref{datailed_training_setting}}.

%employing a sigmoid loss function 

%We adopt the Online Filter strategy proposed in MM-Eureka~\cite{meng2025mmeurekaexploringfrontiersmultimodal}, which filters out all-zero and all-one samples to exclude them from gradient updates.

% Meanwhile, our DPO algorithm is implemented using the trl framework. In this stage, we set global training batch size to 32, learning rate to 5e-7 and dpo beta to 0.1.

\textbf{Training Dataset}.
Our experiments involve multiple training stages (Table~\ref{tab:training_stage}). The first stage mainly activated model reasoning using data from long documents (QuALITY~\cite{pang-etal-2022-quality}), text mathematics (DAPO-Math~\cite{yu2025dapoopensourcellmreinforcement}), and image reasoning (ViRL-39K~\cite{vl-rethinker}). The second stage focuses on image and video data. To mitigate the scarcity of high-quality video data, a filtered subset of diverse video benchmarks, carefully checked for leakage with evaluation datasets, is incorporated. Image data includes subsets from ViRL-39K (Science-Image, Spatial-Image), SPAR-Bench~\cite{zhang2025from} (Spatial-Image), and MME-RealWorld~\cite{zhang2024mme} (General-Image). Video data utilizes several benchmarks: MVBench~\cite{li2023mvbench}, TempCompass~\cite{liu2024tempcompass}, LongVideoBench~\cite{wu2024longvideobench}, HourVideo~\cite{chandrasegaran2024hourvideo}, MLVU~\cite{MLVU}, STI-Bench~\cite{li2025sti}, and VideoVista-CulturalLingo~\cite{chen2025videovistaculturallingo}, along with a filtered 5K data of LLaVA-Video-178K \cite{zhang2024videoinstructiontuningsynthetic}.

\begin{table}
  \caption{Model performance on video reasoning and long video understanding benchmarks. Models with grey backgrounds have $>$11B parameters; those with green backgrounds are based on Qwen2.5-VL-7B. \textcolor[rgb]{ .753,  0,  0}{\textbf{Bold}} values indicate the best performance, and \textcolor[rgb]{ 0,  .439,  .753}{\underline{underlined}} values indicate the second best.}
  \label{tab:main_results}
  \centering
  \renewcommand{\arraystretch}{1.0}
  \resizebox{\textwidth}{!}{%
  \begin{tabular}{llcccccc}
    \toprule
    \multirow{2}{*}{Model} & \multirow{2}{*}{Params} & \multicolumn{4}{c}{Video Reasoning} & \multicolumn{2}{c}{Long Video Understanding} \\
    \cmidrule(lr){3-6} \cmidrule(lr){7-8}
    & & VSI-Bench & Video-MMMU & MMVU (mc) & TOMATO & LVBench & Video-MME (w/o sub) \\
    \midrule
    \rowcolor{gray!15}
     GPT-4o  \cite{openai2024gpt4ocard}          &  -  &  34.0 & 61.2 & -     & 37.7 & 48.9 & 71.9 \\
    \rowcolor{gray!15}
     Gemini 1.5 pro \cite{geminiteam2024gemini15unlockingmultimodal}   &   -  & 45.4  & 53.8  &  -  & 36.1 & 33.1 & 75.0 \\
    \midrule
    mPLUG-Owl3 \cite{ye2024mplugowl3longimagesequenceunderstanding}  & 7B  & -    & 42.0 & - & - & 43.5 & 53.5 \\
    LongVA \cite{zhang2024longcontexttransferlanguage}                    & 7B  & 29.2 & 23.9 & -     & -  & - & 52.6 \\
    LLaVA-Video \cite{zhang2024videoinstructiontuningsynthetic}               & 7B  & 35.6 & 36.1 & - & - & - & 63.3   \\
    % LLaVA-Video                & 72B  & 40.9 & 59.7 & - & - & - & 70.5 \\
    LLaVA-OneVision \cite{li2024llavaonevisioneasyvisualtask}           & 7B  & 32.4 & 33.8 & 49.2 &  - & - & 58.2 \\
    VideoLLaMA2  \cite{cheng2024videollama2advancingspatialtemporal}            & 7B  & -    & -    & 44.8 & -  & - & 47.9 \\
    VideoLLaMA3  \cite{zhang2025videollama3frontiermultimodal}    & 7B  &-  & 47.0  &- & - & 45.3  &66.2 \\
    VILA-1.5  \cite{lin2024vilapretrainingvisuallanguage}                 & 8B  & 28.9 & 20.8 & - & - & - & - \\
    \rowcolor{gray!15}
    VILA-1.5  \cite{lin2024vilapretrainingvisuallanguage}                 & 40B & 31.2 & 34.0 & -     &  - & - & 60.1 \\
    InternVL2  \cite{opengv2024intern20}                & 8B  & 34.6 & 37.4 & 39.0   & 21.7 & - & 54.0 \\
    \rowcolor{gray!15}
    InternVL2   \cite{opengv2024intern20}               & 40B & 36.0 & -    & -    & 29.0  & 39.6 & 61.2 \\
    InternVL2.5   \cite{chen2025expandingperformanceboundariesopensource}             & 8B  & -    & -    & - & -  & - & 64.2 \\
    \rowcolor{gray!15}
    InternVL2.5   \cite{chen2025expandingperformanceboundariesopensource}             & 26B & -    & -    & - & -  & - & 66.9  \\
    %InternVL2.5                & 40B & -    & -    & - & -  & - & 70.7 \\
    InternVideo2.5  \cite{wang2025internvideo25empoweringvideomllms}            & 8B  & -    & 43.0 & - & - &  \textcolor[rgb]{ .753,  0,  0}{\textbf{46.4}} & 65.1 \\
    \rowcolor{gray!15}
    Llama-3.2-Vision \cite{meta2024llama32}  & 11B & 20.6 & 41.8 & - & 21.5 & - & 46.0  \\
    \rowcolor{gray!15}
    Gemma-3-IT    \cite{gemmateam2025gemma3technicalreport} & 12B & 32.4 & \textcolor[rgb]{ 0,  .439,  .753}{\underline{57.2}} & - & 28.1 & - & 58.2  \\
    \rowcolor{gray!15}
    Kimi-VL   \cite{kimiteam2025kimivltechnicalreport}         & 16B (A3B) & 37.4 & 52.6 & - & \textcolor[rgb]{ 0,  .439,  .753}{\underline{31.7}} & -    & \textcolor[rgb]{ .753,  0,  0}{\textbf{67.8}} \\
    \rowcolor{gray!15}
    DeepSeek-VL2  \cite{wu2024deepseekvl2mixtureofexpertsvisionlanguagemodels}     & 28B (A4B) & 21.7 & - & - & 27.2 & - & - \\
    % Qwen2.5-VL(w/o.t.)         & 3B  & 31.4 & 47.4 & 63.5 & 28.8 & 39.0 & 61.7 \\
    
    % Qwen2.5-VL (w/o.t.)         & 7B  & 39.2 & 54.3 & 67.2 & 29.3 & 45.3 & 65.1 \\
    % \rowcolor{gray!15}
    % Qwen2.5-VL (w/o.t.)         & 32B & 40.1 & 62.1 & 70.9 & 33.9 & 44.5 & 70.2 \\
    \rowcolor{green!15}
    Qwen2.5-VL   \cite{bai2025qwen25vltechnicalreport}       & 7B  & 37.5 & 54.3 & \textcolor[rgb]{ .753,  0,  0}{\textbf{67.2}} & 29.3 &  \textcolor[rgb]{ 0,  .439,  .753}{\underline{42.8}} & 66.2 \\
    % \rowcolor{gray!15}
    % Qwen2.5-VL   \cite{bai2025qwen25vltechnicalreport}     & 32B & 40.1 & 62.1 & 70.9 & 33.9 & 44.5 & 70.2 \\
    %\rowcolor{gray!15}   %\cellcolor{blue!20}
    \midrule
    % Qwen2.5-VL (w.t.)           & 3B  & 14.7 & 35.9 & 54.2 & 16.7 & 35.2 & 55.2 \\
    TinyLLaVA-Video-R1\cite{zhang2025tinyllavavideor1smallerlmmsvideo} & 3B & - & - & 46.9 & - & - & 46.6 \\
    \rowcolor{green!15}
    Qwen2.5-VL (thinking)  \cite{bai2025qwen25vltechnicalreport}         & 7B  & 23.8 & 46.8 & 63.0 & 25.8 & 35.2 & 60.4 \\
    \rowcolor{green!15}
    Video-R1   \cite{feng2025videor1reinforcingvideoreasoning}          & 7B  & 35.8 & 52.3 & 64.3 & -     & -     & 59.3\\
    %Qwen2.5-VL (w.t.)           & 32B & 34.1 & - & 58.9 & \\
    Kimi-VL-Thinking \cite{kimiteam2025kimivltechnicalreport} & 16B (A3B) & 32.2 & - & 56.8 & 20.6 & 30.0 & -\\
    
    % \midrule
    %VerIPO (Reasoning Activation) & 7B  & 38.7 & 56.7 & 65.8 & 28.3 & 39.9 & 65.9 \\
    % VerIPO (GRPO-Iteration1)      & 7B  & 41.9 & 56.9 & 66.6 & 31.4 & 41.5 & 66.7 \\
    % VerIPO (DPO-Iteration1)       & 7B  & 41.2 & 56.0 & 65.9 & 31.6 & 41.5 & 67.0 \\
    \rowcolor{green!15}
    VerIPO (Iteration1)       & 7B  & \textcolor[rgb]{ .753,  0,  0}{\textbf{41.8}} & 56.2 & 65.9 & 31.6 & 41.5 & 67.2 \\
    % VerIPO (GRPO-Iteration2)      & 7B & 41.1 & 56.7 & 65.4 & 32.7 & 41.7 & 67.2 \\
    % VerIPO (DPO-Iteration2)       & 7B  & 40.3 & 57.9 & 66.9 & 31.5 & 41.1 & 67.6 \\
    \rowcolor{green!15}
    VerIPO (Iteration2)        & 7B  & 41.0 & \textcolor[rgb]{ .753,  0,  0}{\textbf{57.9}} & \textcolor[rgb]{ 0,  .439,  .753}{\underline{66.9}} & 31.5 & 41.7 & \textcolor[rgb]{ 0,  .439,  .753}{\underline{67.6}} \\
    \rowcolor{green!15}
    % VerIPO (GRPO-Iteration3)      & 7B & 40.9 & 57.0 &  65.4 & 31.9 & & \\
    VerIPO (Iteration3)       & 7B & \textcolor[rgb]{ 0,  .439,  .753}{\underline{41.3}} & 56.8 &  66.7 & \textcolor[rgb]{ .753,  0,  0}{\textbf{32.2}} & 41.7 & 67.2 \\
    \bottomrule
  \end{tabular}
  }
  \vspace{-14pt}
\end{table}

\textbf{Benchmark}.
We introduce four video reasoning benchmarks: VSI-Bench~\cite{yang2024think}, TOMATO~\cite{shangguan2024tomatoassessingvisualtemporal}, Video-MMMU~\cite{hu2025videommmuevaluatingknowledgeacquisition}, MMVU~\cite{zhao2025mmvu} and two long video understanding benchmarks: LVBench~\cite{wang2024lvbench}, and Video-MME~\cite{fu2024videomme}. 
Specifically, VSI-Bench evaluates spatial reasoning, TOMATO assesses temporal reasoning, and Video-MMMU/MMVU tests domain-specific knowledge from multi-discipline videos. LVBench and Video-MME are general benchmarks for comprehensive long video understanding. The detailed evaluation setting of our experiment is in  \textcolor{blue}{\ref{detailed_evaluation_setting}} and evaluation prompt is in  \textcolor{blue}{\ref{cot_prompt}}

%VSI-Bench evaluates a model’s capacity for spatial reasoning in video contexts, while TOMATO focuses on temporal reasoning capabilities. Video-MMMU and MMVU assess a model’s ability to acquire and apply domain-specific knowledge from videos spanning multiple academic disciplines. LVBench and Video-MME are general-purpose benchmarks for long video understanding, providing a comprehensive evaluation of overall video comprehension.

\vspace{-2pt}

\subsection{Results and Analysis}

\textbf{Main Results}. In Table~\ref{tab:main_results}, we present a comparison between our VerIPO iteration and several baseline models, including Qwen2.5-VL and Kimi-VL, across six evaluation benchmarks. It can be observed that VerIPO demonstrates outstanding performance on the video reasoning benchmarks VSI-Bench, Video-MMMU, and TOMATO compared to direct inference and powerful thinking models. There is also a slight improvement in model performance on the long-video general evaluation LVBench ($>$30 minutes) and Video-MME. For complex reasoning, e.g., Video-MMMU and VSI-Bench, we can see large performance increases compared to direct-inference and thinking models, e.g., $\uparrow$ 5.6\% than Video-R1 on Video-MMMU.

% \subsubsection{Ablation Study}

\begin{table}
  \caption{Accuracy and length analysis during training. ``\textbf{Acc-True}'' indicates the answer is correct and also consistent with the right reasoning process. ``Length'' refers to the average token number of responses. The value of Acc-True shows a continual improvement with iterative policy refinement.}
  \label{tab:dpo_experiment}
  \centering
  \renewcommand{\arraystretch}{1.0}
  \resizebox{0.95\textwidth}{!}{%
  \begin{tabular}{lccccccccc}
    \toprule
    Model & \multicolumn{2}{c}{VSI-Bench} & \multicolumn{2}{c}{Video-MMMU} & \multicolumn{2}{c}{MMVU (mc)} & \multicolumn{2}{c}{TOMATO} \\
    \cmidrule{2-9}
    & Acc-True & Length & Acc-True & Length & Acc-True & Length & Acc-True & Length \\
    \midrule
    \rowcolor{gray!15}
    VerIPO (GRPO-Iteration1) &  38.4 & 132 & 51.2 & 308 & 60.7 & 154 & 26.5 & 89 \\
    \midrule
    + \textit{Training with Video-Image Data} \\
    VerIPO (GRPO-Iteration2 w/o DPO) &  36.7 & 138 & 52.3 & 306 & 60.5 & 157 & 25.7 & 91 \\
    VerIPO (DPO-Iteration1) & 38.4 & 167 & 52.9 & 388 & 63.6 & 249 & 28.9 & 161\\ 
    VerIPO (GRPO-Iteration2)  & 37.8 & 164  & 52.2 & 378 &  61.3 & 253 & 27.3 & 167\\ 
    VerIPO (DPO-Iteration2) & 38.4 & 181 & 53.5 & 413 & 65.3 & 275 & 29.2 & 196\\      
    \midrule
    + \textit{Training with Video-Only Data} \\
    VerIPO (DPO-Iteration1) & 38.7 & 167 & 52.0 & 353 & 63.6 & 201 & 28.7 & 113 \\
    VerIPO (GRPO-Iteration2) & 38.3 & 163 & 53.3 & 341 & 61.6 & 194 & 29.1 & 108 \\
    VerIPO (DPO-Iteration2) & 39.4 & 183 & \textcolor[rgb]{ .753,  0,  0}{\textbf{56.7}} & 382 & \textcolor[rgb]{ .753,  0,  0}{\textbf{66.1}} & 243 & 31.0 & 134 \\
    VerIPO (GRPO-Iteration3) & 39.9  & 172 & 55.3 & 359 & 64.3 & 223 & 31.0 & 146 \\
    VerIPO (DPO-Iteration3) & \textcolor[rgb]{ .753,  0,  0}{\textbf{40.1}} & \textcolor[rgb]{ .753,  0,  0}{\textbf{200}}  & 55.4  & \textcolor[rgb]{ .753,  0,  0}{\textbf{430}}  & 65.9 & \textcolor[rgb]{ .753,  0,  0}{\textbf{291}} & \textcolor[rgb]{ .753,  0,  0}{\textbf{31.3}} & \textcolor[rgb]{ .753,  0,  0}{\textbf{202}} \\
    \bottomrule
  \end{tabular} 
  }
  \vspace{-7pt}
\end{table}

\textbf{Effects of Verifier-Guided DPO}.
Table~\ref{tab:dpo_experiment} illustrates iterative policy optimization with different data and strategies. All models use identical video/image data for the initial GRPO round (GRPO-Iteration1). Verifier-guided DPO improves true accuracy and thinking length compared to GRPO. The continual GRPO does not bring an increase in accuracy and low length. Subsequent iterations, trained solely on video data after initial reasoning activation, show stable accuracy improvement. 

%To maintain the consistency of continuous GRPO iterations, the ablation experiment uses the origin dataset of images and videos, rather than the video-only data used in the previous VerIPO iterations. 

% between two rounds of GRPO iterations, one round of VerIPO iterations, and two rounds of VerIPO iterations.
% Among them, the two rounds of GRPO iterations(1 round of GRPO + 1 round of GRPO) are similar to one round of VerIPO iterations(1 round of GRPO + 1 round of DPO), as both involve training using data twice. To maintain the consistency of continuous GRPO iterations, this experiment uses the origin dataset of images and videos, rather than the video-only data used in the previous VerIPO iterations. 
% From the experimental results, we can observe that one round of VerIPO (1 round of GRPO + 1 round of DPO) outperforms the two rounds of GRPO in both accuracy and response length. Furthermore, as the number of VerIPO iterations increases, the performance continues to improve.

\begin{figure}
    \centering
    \includegraphics[width=0.95\linewidth]{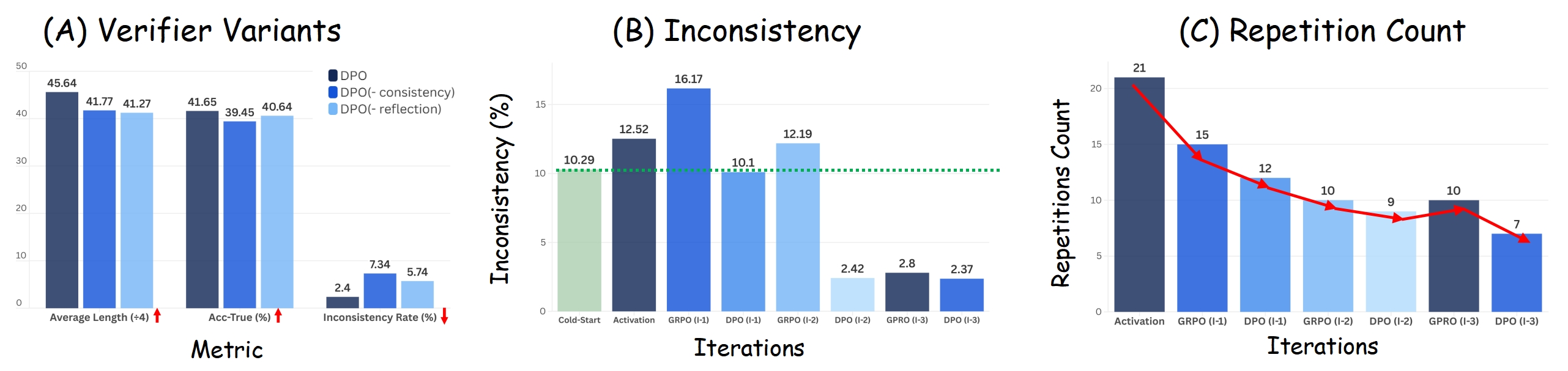}
    \caption{Figure (A): Performance comparison after removing \textit{Reflective Preference Pairs} and \textit{Inference Consistency Pairs} during DPO (I-2) stage. The reported values represent the average metric across the MMVU (mc) and TOMATO. For visualization, the response length has been scaled down to 0.25 of the original. Figure (B): Inconsistency rate (thinking vs. final answer) at Cold Start and different stages of VerIPO. The reported values represent the average scores across the MMVU (mc) and TOMATO. The statistical inconsistency rate is in \ref{inconsistency_rate}. Figure (C): The number of repeated responses generated by VerIPO at different training stages over the evaluation datasets. The reported values are computed as the sum of VSI-Bench, Video-MMMU, MMVU (mc) and TOMATO.}
    \label{fig:experiment}
    \vspace{-15pt}
\end{figure}

\textbf{Impacts of Verifier}.  Figure \ref{fig:experiment} (A) illustrates the impact of different DPO preference pairs. Statistical analysis shows that training without consistency or reflective pairs leads to decreased response length and accuracy, particularly hindering improvements in the inconsistency rate.

\textbf{Iterations}.
An ablation study on the number of VerIPO iterations is shown in Table~\ref{tab:dpo_experiment} (last five rows). We observe that increasing iterations leads to improved reasoning consistency (Figure \ref{fig:absfigure}), true accuracy (26.5 $\rightarrow$ 31.3 in TOMATO), and response length (average 100 tokens increase). Overall, the iteration of VerIPO can continually improve the reasoning length (mainly driven by fast DPO training) and improve the true accuracy (removing ``correct answer with error thinking'').

\begin{table}[t]
  \caption{Performance comparison across different training methods (SFT or reasoning activation)}
  \label{tab:abalation_study}
  \centering
  \renewcommand{\arraystretch}{1.0}
  \resizebox{0.95\textwidth}{!}{%
  \begin{tabular}{llcccccc}
    \toprule
   Stage & Method & VSI-Bench & Video-MMMU & MMVU (mc) & TOMATO & LVBench & Video-MME(w/o sub) \\
    \midrule
    \multirow{2}*{Start}& Qwen2.5-VL-7B (w/o.t.) & 37.5 & 54.3 & 67.2 & 29.3 &  \textcolor[rgb]{ .753,  0,  0}{\textbf{42.8}} & 66.2\\
    & Qwen2.5-VL-7B (w.t.) & 23.8 & 46.8 & 63.0 & 25.8 & 35.2 & 60.4\\
   \multirow{2}*{Warm} & \textcolor{red}{+ SFT} & 33.8 & 53.4 & 65.8 & 26.8 & 36.1 & 57.8  \\
    & \textcolor{blue}{+ Reasoning Activation} & 38.7 & 56.7 & 65.8 & 28.3 & 39.9 & 65.9 \\
    \midrule
   \multirow{3}*{Iter-1} & + GRPO & 33.4 & 54.0 & 66.1 & 28.6 & 40.7 & 64.7 \\
    & \textcolor{red}{+ GRPO (w SFT)} & 36.6 & 55.9 & 65.3 & 29.8 & 38.4 & 61.0 \\
    & \textcolor{blue}{+ GRPO (w activation)} &  \textcolor[rgb]{ .753,  0,  0}{\textbf{41.9}} & 56.9 & 66.6 & 31.4 & 41.5 & 66.7 \\   
    \midrule
    \multirow{3}*{Iter-1}&+ DPO & 33.9 & 54.2 & 66.9 & 28.2 & 40.3 & 64.2\\
    &\textcolor{red}{+ DPO (w SFT)} & 36.6 & 53.8 & 66.2 & 28.7 & 35.8 & 60.0 \\
    &\textcolor{blue}{+ DPO (w activation)} & 41.8 & 56.2 & 65.9 & 31.6 & 41.5 &  \textcolor[rgb]{ .753,  0,  0}{\textbf{67.2}} \\ 
    \midrule
    \multirow{3}*{Iter-2}&+ GRPO & 36.0 & 52.9 & \textcolor[rgb]{ .753,  0,  0}{\textbf{68.0}} & 30.1 & 41.0 & 65.8 \\
    &\textcolor{red}{+ GRPO (w SFT)} & 37.0 & 54.0 & 66.7 & 28.5 & 39.3 & 62.8 \\
    &\textcolor{blue}{+ GRPO (w activation)} & 41.3 &  \textcolor[rgb]{ .753,  0,  0}{\textbf{56.9}}  & 65.4 & \textcolor[rgb]{ .753,  0,  0}{\textbf{32.7}} & 41.7 & 67.0\\
    \bottomrule
  \end{tabular} 
  }
  \vspace{-4pt}
\end{table}

\textbf{Impact of Cold Start}.
We evaluate Cold Start (\textcolor{red}{SFT}) in RFT using the Video-R1-COT 165k dataset. Table~\ref{tab:abalation_study} reveals that Cold Start training resulted in marginal gains on metrics like Video-MMMU but substantial performance degradation on general reasoning tasks, unrecoverable by subsequent VerIPO iterations. This highlights the impact of low-quality video Cold Start data on performance in video tasks. However, VerIPO (iteration with only video data (black) or \textcolor{blue}{activation}), starting from RL not SFT, show more stable performance improvement across benchmarks.

\textbf{Reasoning Activation}. 
In Table~\ref{tab:abalation_study}, we also present a comparative analysis of the experimental results with and without Reasoning Activation. It is observed that the reasoning activation phase, which only utilizes image and text data, demonstrates good generalization in the video domain, particularly exhibiting exceptionally strong performance on the Video-MMMU (complex domain reasoning) task.

\begin{figure}[t]
    \centering
    \includegraphics[width=0.95\linewidth]{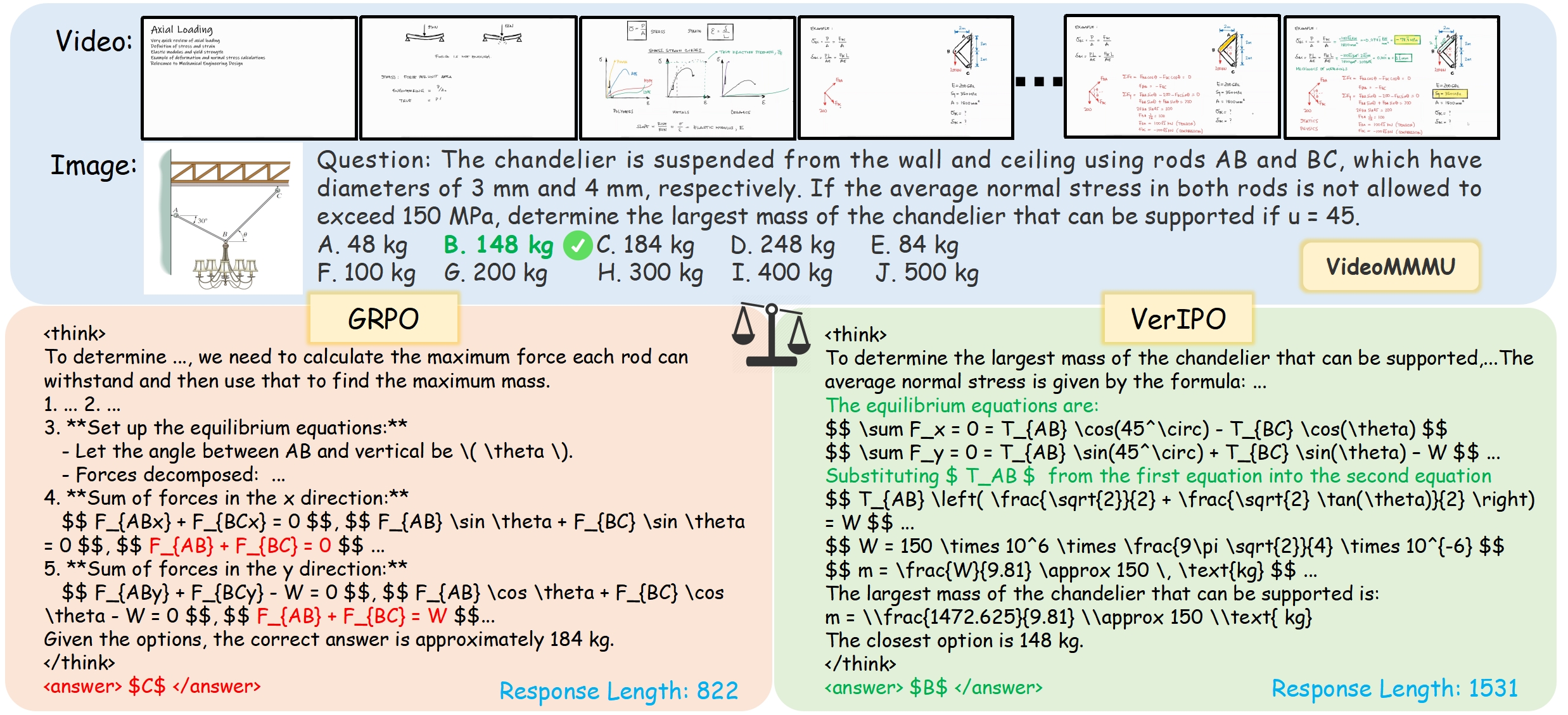}
    \caption{A case from Video-MMMU shows the comparative performance of GRPO and VerIPO. Our method can generate longer CoTs with accurate and logical formulas to solve physical problems.}
    \label{fig:case}
    \vspace{-10pt}
\end{figure}

\textbf{Inconsistency of reasoning process and answer}.
The experiments reveal inconsistency between the reasoning process and final answer, also observed in the cold start (SFT) experiments (as shown in Figure \ref{fig:experiment} (B)). The implementation of our VerIPO training loop successfully addresses this issue, demonstrating a progressive reduction in inconsistency across training iterations and a consistent enhancement in true accuracy (as shown in Table \ref{tab:dpo_experiment}).

\textbf{Repetition of generated content}.
 Repetitive inference loops exceeding context limits were observed in our experiments and with Kimi-VL-Thinking evaluation. 
 Using the Rollout-Aware Verifier to construct specialized preference data addressed this issue, with DPO iterations significantly decreasing repetition frequency (as shown in Figure \ref{fig:experiment} (C)). Moreover, although these large multimodal models demonstrate strong reasoning capabilities with challenging static images and short video clips (limited frames inputting), their performance degrades significantly when processing longer video sequences or several frames inputting. This degradation manifests as repetitive or inconsistent reasoning, and in some cases, a complete inability to reason. We hypothesize this limitation stems from the models' inherent capacity for long-video understanding during pre-training, highlighting a crucial area for future model improvement.

\subsection{Case Study}

% We present a comparative case in Figure \ref{fig:case}, where \textcolor{red}{red} and \textcolor[rgb]{0, 0.6, 0}{green} words represent accurate and error reasoning contents. More cases are shown in Figures xxx-xxx. Based on the above experimental tables and cases, we can see that VerIPO can make models generate longer chains for challenging science problems compared to general reasoning problems and the math data used in reasoning activation may help the logical reasoning activation of science problems. 

Based on Figure \ref{fig:case} (with additional cases in \textcolor{blue}{\ref{qualitativa_analysis}}), where \textcolor{red}{red} indicates error reasoning and \textcolor[rgb]{0, 0.6, 0}{green} accurate reasoning, and previous experimental analysis, we observe VerIPO enables models to generate longer and more accurate reasoning chains (sometimes with reflection) for challenging science, temporal grounding problems besides general reasoning tasks, compared to GRPO training. 
In addition, we observe that utilizing textual or visual math in the reasoning activation stage may aid logical reasoning, based on the reasoning process of GRPO and VerIPO in science problems.

\subsection{Discussion}

\textbf{Why do RL-trained LMRMs struggle to achieve consistent performance increase in all Video tasks?} 
\textit{1) High-Quality and Diverse Video Reasoning Data (Verifiable Data)}: Training LMRMs with RL requires vast amounts of high-quality data, particularly for video reasoning tasks that demand strong reasoning abilities or involve long reasoning paths. Most existing video datasets are primarily focused on simple recognition or short-term actions, lacking the complexity and scale needed for robust RL training.
\textit{2) Model Capability Limitations in Video Understanding (Foundation Models)}: The base model, upon which LMRMs are built, often relies on pre-training methodologies that are not ideally suited for comprehensive video understanding, especially over long durations. While these foundation models excel at learning powerful representations from vast amounts of image-text pairs or short video clips, their pre-training objectives typically do not fully capture the nuances of long-range temporal dependencies, event causality and sequence, and contextual consistency over time.
\textit{3) Cold Start Problem (Data Quality)}: If RL is used for fine-tuning after a supervised fine-tuning (SFT) phase, a poor initial SFT policy (especially for video) can hinder the RL agent's ability to explore effectively and find optimal policies.

\textbf{Why do direct-answer models outperform long-thinking model variants?}
\textit{1) Instability and sensitivity of RL training}: The inherent instability of RL can make ``long-thinking" approaches particularly challenging to optimize for long visual inputs (video). RL training for long-thinking models is hampered by their expansive "action space", which makes efficient exploration difficult and can lead to getting stuck in suboptimal solutions. This complexity also exacerbates hyperparameter sensitivity, a common RL challenge, risking training instability. Direct-answer models benefit from a smaller output space, simplifying both exploration.
\textit{2) Not all prompts require thinking (\underline{Overthinking})}: The benefit of ``long-thinking" is task-dependent. For many common prompts, a direct answer is sufficient, and forcing a reasoning process can introduce unnecessary complexity, computational overhead, and potential thinking errors. We should build LMRMs to perform adaptive reasoning for different prompts.
\textit{3) RL data size is limited}: The effectiveness of RL, especially for complex generative tasks, is highly dependent on the quantity and quality of data. The limitations in RL data directly impact the ability of long-thinking models to learn effectively.

\textbf{How to build an LMRM with adaptive reasoning capability?}
\textit{1) Reasoning activation for different thinking patterns}: The reasoning activation stage should use diverse data, including direct-answer examples for conciseness, step-by-step reasoning examples for detailed thought processes, mixed modality reasoning to handle various input types, and reasoning-on-demand examples that prompt specific output styles. This multifaceted reasoning activation exposes the model to a range of reasoning strategies, preventing it from being confined to a single, rigid approach.
\textit{2) Reward function for adaptive reasoning}: Effective RL fine-tuning for adaptive reasoning necessitates sophisticated reward functions beyond answer and format accuracy, e.g., including short, middle, or long thinking judge for different prompts. These should include composite rewards that value reasoning quality, conciseness, and coherence; efficiency-aware rewards that penalize overthinking on simple problems; and adaptive policy rewards that dynamically adjust based on problem complexity. Such nuanced signals guide the model to select the appropriate depth and style of reasoning for different prompts.
\textit{3) Iterative optimization enhancement strategy}: The most effective development of adaptive reasoning in an LMRM may occur through an iterative optimization loop. This loop strategically blends enforced SFT, target-optimization DPO, and wide exploration GRPO, collectively allowing the model to progressively refine its capacity for selecting and executing the optimal reasoning strategy tailored to various video understanding tasks.

\section{Conclusion}

 Addressing the challenge of deep reasoning in Video-LLMs, we propose \textbf{VerIPO}, a novel online rollout-aware $\textbf{Ver}$ifier-guided $\textbf{I}$terative $\textbf{P}$olicy $\textbf{O}$ptimization algorithm. This RL-based GRPO-Verifier-DPO loop employs a small LLM verifier to refine generated CoTs, cultivating reasoning capability efficiently without requiring large Long-CoT SFT cold starts. VerIPO significantly improves reasoning consistency, accuracy, and response length, outperforming larger and more powerful baselines on video benchmarks. 
 While effective, \textit{limitations} include potential verifier dependence, limited data size and response length, and computational costs. \textit{Future works} aims to address these by exploring verifier designs, optimizing the pipeline and leveraging GRPO exploration, targeted DPO, and strong SFT on base LMMs towards achieving robust, long-term reasoning ability across unimodality.

\bibliographystyle{ieee_fullname}
\bibliography{ref}

\begin{thebibliography}{10}\itemsep=-1pt

\bibitem{arnab2021vivitvideovisiontransformer}
Anurag Arnab, Mostafa Dehghani, Georg Heigold, Chen Sun, Mario Lučić, and Cordelia Schmid.
\newblock Vivit: A video vision transformer, 2021.

\bibitem{bai2025qwen25vltechnicalreport}
Shuai Bai, Keqin Chen, Xuejing Liu, Jialin Wang, Wenbin Ge, Sibo Song, Kai Dang, Peng Wang, Shijie Wang, Jun Tang, Humen Zhong, Yuanzhi Zhu, Mingkun Yang, Zhaohai Li, Jianqiang Wan, Pengfei Wang, Wei Ding, Zheren Fu, Yiheng Xu, Jiabo Ye, Xi Zhang, Tianbao Xie, Zesen Cheng, Hang Zhang, Zhibo Yang, Haiyang Xu, and Junyang Lin.
\newblock Qwen2.5-vl technical report, 2025.

\bibitem{chandrasegaran2024hourvideo}
Keshigeyan Chandrasegaran, Agrim Gupta, Lea~M. Hadzic, Taran Kota, Jimming He, Cristobal Eyzaguirre, Zane Durante, Manling Li, Jiajun Wu, and Fei-Fei Li.
\newblock Hourvideo: 1-hour video-language understanding.
\newblock In {\em Advances in Neural Information Processing Systems}, volume~37, 2024.

\bibitem{chen2024mecdunlockingmultieventcausal}
Tieyuan Chen, Huabin Liu, Tianyao He, Yihang Chen, Chaofan Gan, Xiao Ma, Cheng Zhong, Yang Zhang, Yingxue Wang, Hui Lin, and Weiyao Lin.
\newblock Mecd: Unlocking multi-event causal discovery in video reasoning, 2024.

\bibitem{chen2025suitabilityreinforcementfinetuningvisual}
Xiaxu Chen, Wei Li, Chunxu Liu, Chi Xie, Xiaoyan Hu, Chengqian Ma, Feng Zhu, and Rui Zhao.
\newblock On the suitability of reinforcement fine-tuning to visual tasks, 2025.

\bibitem{chen2025videovistaculturallingo}
Xinyu Chen, Yunxin Li, Haoyuan Shi, Baotian Hu, Wenhan Luo, Yaowei Wang, and Min Zhang.
\newblock Videovista-culturallingo: 360$^\circ$ horizons-bridging cultures, languages, and domains in video comprehension, 2025.

\bibitem{chen2025visrlintentiondrivenvisualperception}
Zhangquan Chen, Xufang Luo, and Dongsheng Li.
\newblock Visrl: Intention-driven visual perception via reinforced reasoning, 2025.

\bibitem{chen2025expandingperformanceboundariesopensource}
Zhe Chen, Weiyun Wang, Yue Cao, Yangzhou Liu, Zhangwei Gao, Erfei Cui, Jinguo Zhu, Shenglong Ye, Hao Tian, Zhaoyang Liu, Lixin Gu, Xuehui Wang, Qingyun Li, Yimin Ren, Zixuan Chen, Jiapeng Luo, Jiahao Wang, Tan Jiang, Bo Wang, Conghui He, Botian Shi, Xingcheng Zhang, Han Lv, Yi Wang, Wenqi Shao, Pei Chu, Zhongying Tu, Tong He, Zhiyong Wu, Huipeng Deng, Jiaye Ge, Kai Chen, Kaipeng Zhang, Limin Wang, Min Dou, Lewei Lu, Xizhou Zhu, Tong Lu, Dahua Lin, Yu Qiao, Jifeng Dai, and Wenhai Wang.
\newblock Expanding performance boundaries of open-source multimodal models with model, data, and test-time scaling, 2025.

\bibitem{cheng2024videollama2advancingspatialtemporal}
Zesen Cheng, Sicong Leng, Hang Zhang, Yifei Xin, Xin Li, Guanzheng Chen, Yongxin Zhu, Wenqi Zhang, Ziyang Luo, Deli Zhao, and Lidong Bing.
\newblock Videollama 2: Advancing spatial-temporal modeling and audio understanding in video-llms, 2024.

\bibitem{chris2025skyworkr1v2multimodalhybrid}
Chris, Yichen Wei, Yi Peng, Xiaokun Wang, Weijie Qiu, Wei Shen, Tianyidan Xie, Jiangbo Pei, Jianhao Zhang, Yunzhuo Hao, Xuchen Song, Yang Liu, and Yahui Zhou.
\newblock Skywork r1v2: Multimodal hybrid reinforcement learning for reasoning, 2025.

\bibitem{dai2023instructblipgeneralpurposevisionlanguagemodels}
Wenliang Dai, Junnan Li, Dongxu Li, Anthony Meng~Huat Tiong, Junqi Zhao, Weisheng Wang, Boyang Li, Pascale Fung, and Steven Hoi.
\newblock Instructblip: Towards general-purpose vision-language models with instruction tuning, 2023.

\bibitem{daxberger2025mmspatialexploring3dspatial}
Erik Daxberger, Nina Wenzel, David Griffiths, Haiming Gang, Justin Lazarow, Gefen Kohavi, Kai Kang, Marcin Eichner, Yinfei Yang, Afshin Dehghan, and Peter Grasch.
\newblock Mm-spatial: Exploring 3d spatial understanding in multimodal llms, 2025.

\bibitem{deepseekai2025deepseekr1incentivizingreasoningcapability}
DeepSeek-AI, Daya Guo, Dejian Yang, Haowei Zhang, et~al.
\newblock Deepseek-r1: Incentivizing reasoning capability in llms via reinforcement learning, 2025.

\bibitem{deng2025boostinggeneralizationreasoningvision}
Huilin Deng, Ding Zou, Rui Ma, Hongchen Luo, Yang Cao, and Yu Kang.
\newblock Boosting the generalization and reasoning of vision language models with curriculum reinforcement learning, 2025.

\bibitem{dong2025insightvexploringlongchainvisual}
Yuhao Dong, Zuyan Liu, Hai-Long Sun, Jingkang Yang, Winston Hu, Yongming Rao, and Ziwei Liu.
\newblock Insight-v: Exploring long-chain visual reasoning with multimodal large language models, 2025.

\bibitem{dosovitskiy2021imageworth16x16words}
Alexey Dosovitskiy, Lucas Beyer, Alexander Kolesnikov, Dirk Weissenborn, Xiaohua Zhai, Thomas Unterthiner, Mostafa Dehghani, Matthias Minderer, Georg Heigold, Sylvain Gelly, Jakob Uszkoreit, and Neil Houlsby.
\newblock An image is worth 16x16 words: Transformers for image recognition at scale, 2021.

\bibitem{fei2024videoofthoughtstepbystepvideoreasoning}
Hao Fei, Shengqiong Wu, Wei Ji, Hanwang Zhang, Meishan Zhang, Mong-Li Lee, and Wynne Hsu.
\newblock Video-of-thought: Step-by-step video reasoning from perception to cognition, 2024.

\bibitem{feng2025videor1reinforcingvideoreasoning}
Kaituo Feng, Kaixiong Gong, Bohao Li, Zonghao Guo, Yibing Wang, Tianshuo Peng, Benyou Wang, and Xiangyu Yue.
\newblock Video-r1: Reinforcing video reasoning in mllms, 2025.

\bibitem{fu2024videomme}
Chaoyou Fu, Yuhan Dai, Yondong Luo, Lei Li, Shuhuai Ren, Renrui Zhang, Zihan Wang, Chenyu Zhou, Yunhang Shen, Mengdan Zhang, et~al.
\newblock Video-mme: The first-ever comprehensive evaluation benchmark of multi-modal llms in video analysis.
\newblock {\em arXiv preprint arXiv:2405.21075}, 2024.

\bibitem{gupta2025ampoactivemultipreferenceoptimization}
Taneesh Gupta, Rahul Madhavan, Xuchao Zhang, Chetan Bansal, and Saravan Rajmohan.
\newblock Ampo: Active multi-preference optimization, 2025.

\bibitem{hu2025videommmuevaluatingknowledgeacquisition}
Kairui Hu, Penghao Wu, Fanyi Pu, Wang Xiao, Yuanhan Zhang, Xiang Yue, Bo Li, and Ziwei Liu.
\newblock Video-mmmu: Evaluating knowledge acquisition from multi-discipline professional videos, 2025.

\bibitem{huang2025visionr1incentivizingreasoningcapability}
Wenxuan Huang, Bohan Jia, Zijie Zhai, Shaosheng Cao, Zheyu Ye, Fei Zhao, Zhe Xu, Yao Hu, and Shaohui Lin.
\newblock Vision-r1: Incentivizing reasoning capability in multimodal large language models, 2025.

\bibitem{hurst2024gpt}
Aaron Hurst, Adam Lerer, Adam~P Goucher, Adam Perelman, Aditya Ramesh, Aidan Clark, AJ Ostrow, Akila Welihinda, Alan Hayes, Alec Radford, et~al.
\newblock Gpt-4o system card.
\newblock {\em arXiv preprint arXiv:2410.21276}, 2024.

\bibitem{li2024llavaonevisioneasyvisualtask}
Bo Li, Yuanhan Zhang, Dong Guo, Renrui Zhang, Feng Li, Hao Zhang, Kaichen Zhang, Peiyuan Zhang, Yanwei Li, Ziwei Liu, and Chunyuan Li.
\newblock Llava-onevision: Easy visual task transfer, 2024.

\bibitem{li2022mplugeffectiveefficientvisionlanguage}
Chenliang Li, Haiyang Xu, Junfeng Tian, Wei Wang, Ming Yan, Bin Bi, Jiabo Ye, Hehong Chen, Guohai Xu, Zheng Cao, Ji Zhang, Songfang Huang, Fei Huang, Jingren Zhou, and Luo Si.
\newblock mplug: Effective and efficient vision-language learning by cross-modal skip-connections, 2022.

\bibitem{li2024videochatchatcentricvideounderstanding}
KunChang Li, Yinan He, Yi Wang, Yizhuo Li, Wenhai Wang, Ping Luo, Yali Wang, Limin Wang, and Yu Qiao.
\newblock Videochat: Chat-centric video understanding, 2024.

\bibitem{li2023mvbench}
Kunchang Li, Yali Wang, Yinan He, Yizhuo Li, Yi Wang, Yi Liu, Zun Wang, Jilan Xu, Guo Chen, Ping Luo, Limin Wang, and Yu Qiao.
\newblock Mvbench: A comprehensive multi-modal video understanding benchmark, 2023.

\bibitem{li2025videochatr1enhancingspatiotemporalperception}
Xinhao Li, Ziang Yan, Desen Meng, Lu Dong, Xiangyu Zeng, Yinan He, Yali Wang, Yu Qiao, Yi Wang, and Limin Wang.
\newblock Videochat-r1: Enhancing spatio-temporal perception via reinforcement fine-tuning, 2025.

\bibitem{10598361}
Yunxin Li, Baotian Hu, Xinyu Chen, Lin Ma, Yong Xu, and Min Zhang.
\newblock Lmeye: An interactive perception network for large language models.
\newblock {\em IEEE Transactions on Multimedia}, 26:10952--10964, 2024.

\bibitem{10887014}
Yunxin Li, Shenyuan Jiang, Baotian Hu, Longyue Wang, Wanqi Zhong, Wenhan Luo, Lin Ma, and Min Zhang.
\newblock Uni-moe: Scaling unified multimodal llms with mixture of experts.
\newblock {\em IEEE Transactions on Pattern Analysis and Machine Intelligence}, 47(5):3424--3439, 2025.

\bibitem{li2025perception}
Yunxin Li, Zhenyu Liu, Zitao Li, Xuanyu Zhang, Zhenran Xu, Xinyu Chen, Haoyuan Shi, Shenyuan Jiang, Xintong Wang, Jifang Wang, et~al.
\newblock Perception, reason, think, and plan: A survey on large multimodal reasoning models.
\newblock {\em arXiv preprint arXiv:2505.04921}, 2025.

\bibitem{li2025sti}
Yun Li, Yiming Zhang, Tao Lin, XiangRui Liu, Wenxiao Cai, Zheng Liu, and Bo Zhao.
\newblock Sti-bench: Are mllms ready for precise spatial-temporal world understanding?
\newblock {\em arXiv preprint arXiv:2503.23765}, 2025.

\bibitem{lin2024vilapretrainingvisuallanguage}
Ji Lin, Hongxu Yin, Wei Ping, Yao Lu, Pavlo Molchanov, Andrew Tao, Huizi Mao, Jan Kautz, Mohammad Shoeybi, and Song Han.
\newblock Vila: On pre-training for visual language models, 2024.

\bibitem{liu2025spatialcotadvancingspatialreasoning}
Yuecheng Liu, Dafeng Chi, Shiguang Wu, Zhanguang Zhang, Yaochen Hu, Lingfeng Zhang, Yingxue Zhang, Shuang Wu, Tongtong Cao, Guowei Huang, Helong Huang, Guangjian Tian, Weichao Qiu, Xingyue Quan, Jianye Hao, and Yuzheng Zhuang.
\newblock Spatialcot: Advancing spatial reasoning through coordinate alignment and chain-of-thought for embodied task planning, 2025.

\bibitem{liu2024tempcompass}
Yuanxin Liu, Shicheng Li, Yi Liu, Yuxiang Wang, Shuhuai Ren, Lei Li, Sishuo Chen, Xu Sun, and Lu Hou.
\newblock Tempcompass: Do video llms really understand videos?
\newblock {\em arXiv preprint arXiv: 2403.00476}, 2024.

\bibitem{liu2025videomindchainofloraagentlong}
Ye Liu, Kevin~Qinghong Lin, Chang~Wen Chen, and Mike~Zheng Shou.
\newblock Videomind: A chain-of-lora agent for long video reasoning, 2025.

\bibitem{liu2025segzeroreasoningchainguidedsegmentation}
Yuqi Liu, Bohao Peng, Zhisheng Zhong, Zihao Yue, Fanbin Lu, Bei Yu, and Jiaya Jia.
\newblock Seg-zero: Reasoning-chain guided segmentation via cognitive reinforcement, 2025.

\bibitem{liu2021videoswintransformer}
Ze Liu, Jia Ning, Yue Cao, Yixuan Wei, Zheng Zhang, Stephen Lin, and Han Hu.
\newblock Video swin transformer, 2021.

\bibitem{liu2025visualrftvisualreinforcementfinetuning}
Ziyu Liu, Zeyi Sun, Yuhang Zang, Xiaoyi Dong, Yuhang Cao, Haodong Duan, Dahua Lin, and Jiaqi Wang.
\newblock Visual-rft: Visual reinforcement fine-tuning, 2025.

\bibitem{liu2025othinkmr1stimulatingmultimodalgeneralized}
Zhiyuan Liu, Yuting Zhang, Feng Liu, Changwang Zhang, Ying Sun, and Jun Wang.
\newblock Othink-mr1: Stimulating multimodal generalized reasoning capabilities via dynamic reinforcement learning, 2025.

\bibitem{luo2025guir1generalistr1style}
Run Luo, Lu Wang, Wanwei He, and Xiaobo Xia.
\newblock Gui-r1 : A generalist r1-style vision-language action model for gui agents, 2025.

\bibitem{maaz2024videochatgptdetailedvideounderstanding}
Muhammad Maaz, Hanoona Rasheed, Salman Khan, and Fahad~Shahbaz Khan.
\newblock Video-chatgpt: Towards detailed video understanding via large vision and language models, 2024.

\bibitem{meng2025mmeurekaexploringfrontiersmultimodal}
Fanqing Meng, Lingxiao Du, Zongkai Liu, Zhixiang Zhou, Quanfeng Lu, Daocheng Fu, Tiancheng Han, Botian Shi, Wenhai Wang, Junjun He, Kaipeng Zhang, Ping Luo, Yu Qiao, Qiaosheng Zhang, and Wenqi Shao.
\newblock Mm-eureka: Exploring the frontiers of multimodal reasoning with rule-based reinforcement learning, 2025.

\bibitem{neimark2021videotransformernetwork}
Daniel Neimark, Omri Bar, Maya Zohar, and Dotan Asselmann.
\newblock Video transformer network, 2021.

\bibitem{oquab2024dinov2learningrobustvisual}
Maxime Oquab, Timothée Darcet, Théo Moutakanni, Huy Vo, Marc Szafraniec, Vasil Khalidov, Pierre Fernandez, Daniel Haziza, Francisco Massa, Alaaeldin El-Nouby, Mahmoud Assran, Nicolas Ballas, Wojciech Galuba, Russell Howes, Po-Yao Huang, Shang-Wen Li, Ishan Misra, Michael Rabbat, Vasu Sharma, Gabriel Synnaeve, Hu Xu, Hervé Jegou, Julien Mairal, Patrick Labatut, Armand Joulin, and Piotr Bojanowski.
\newblock Dinov2: Learning robust visual features without supervision, 2024.

\bibitem{ouyang2025spatialr1enhancingmllmsvideo}
Kun Ouyang.
\newblock Spatial-r1: Enhancing mllms in video spatial reasoning, 2025.

\bibitem{pang-etal-2022-quality}
Richard~Yuanzhe Pang, Alicia Parrish, Nitish Joshi, Nikita Nangia, Jason Phang, Angelica Chen, Vishakh Padmakumar, Johnny Ma, Jana Thompson, He He, and Samuel Bowman.
\newblock {Q}u{ALITY}: Question answering with long input texts, yes!
\newblock In {\em Proceedings of the 2022 Conference of the North American Chapter of the Association for Computational Linguistics: Human Language Technologies}, pages 5336--5358, Seattle, United States, July 2022. Association for Computational Linguistics.

\bibitem{peng2025lmmr1empowering3blmms}
Yingzhe Peng, Gongrui Zhang, Miaosen Zhang, Zhiyuan You, Jie Liu, Qipeng Zhu, Kai Yang, Xingzhong Xu, Xin Geng, and Xu Yang.
\newblock Lmm-r1: Empowering 3b lmms with strong reasoning abilities through two-stage rule-based rl, 2025.

\bibitem{radford2021learningtransferablevisualmodels}
Alec Radford, Jong~Wook Kim, Chris Hallacy, Aditya Ramesh, Gabriel Goh, Sandhini Agarwal, Girish Sastry, Amanda Askell, Pamela Mishkin, Jack Clark, Gretchen Krueger, and Ilya Sutskever.
\newblock Learning transferable visual models from natural language supervision, 2021.

\bibitem{rafailov2024directpreferenceoptimizationlanguage}
Rafael Rafailov, Archit Sharma, Eric Mitchell, Stefano Ermon, Christopher~D. Manning, and Chelsea Finn.
\newblock Direct preference optimization: Your language model is secretly a reward model, 2024.

\bibitem{ray2025satdynamicspatialaptitude}
Arijit Ray, Jiafei Duan, Ellis Brown, Reuben Tan, Dina Bashkirova, Rose Hendrix, Kiana Ehsani, Aniruddha Kembhavi, Bryan~A. Plummer, Ranjay Krishna, Kuo-Hao Zeng, and Kate Saenko.
\newblock Sat: Dynamic spatial aptitude training for multimodal language models, 2025.

\bibitem{schulman2017proximalpolicyoptimizationalgorithms}
John Schulman, Filip Wolski, Prafulla Dhariwal, Alec Radford, and Oleg Klimov.
\newblock Proximal policy optimization algorithms, 2017.

\bibitem{shangguan2024tomatoassessingvisualtemporal}
Ziyao Shangguan, Chuhan Li, Yuxuan Ding, Yanan Zheng, Yilun Zhao, Tesca Fitzgerald, and Arman Cohan.
\newblock Tomato: Assessing visual temporal reasoning capabilities in multimodal foundation models, 2024.

\bibitem{shao2024deepseekmathpushinglimitsmathematical}
Zhihong Shao, Peiyi Wang, Qihao Zhu, Runxin Xu, Junxiao Song, Xiao Bi, Haowei Zhang, Mingchuan Zhang, Y.~K. Li, Y. Wu, and Daya Guo.
\newblock Deepseekmath: Pushing the limits of mathematical reasoning in open language models, 2024.

\bibitem{shi2025efficientreinforcementfinetuningadaptive}
Taiwei Shi, Yiyang Wu, Linxin Song, Tianyi Zhou, and Jieyu Zhao.
\newblock Efficient reinforcement finetuning via adaptive curriculum learning, 2025.

\bibitem{sun2025mmverifyenhancingmultimodalreasoning}
Linzhuang Sun, Hao Liang, Jingxuan Wei, Bihui Yu, Tianpeng Li, Fan Yang, Zenan Zhou, and Wentao Zhang.
\newblock Mm-verify: Enhancing multimodal reasoning with chain-of-thought verification, 2025.

\bibitem{tan2025reasonrftreinforcementfinetuningvisual}
Huajie Tan, Yuheng Ji, Xiaoshuai Hao, Minglan Lin, Pengwei Wang, Zhongyuan Wang, and Shanghang Zhang.
\newblock Reason-rft: Reinforcement fine-tuning for visual reasoning, 2025.

\bibitem{tang2025gametheoreticregularizedselfplayalignment}
Xiaohang Tang, Sangwoong Yoon, Seongho Son, Huizhuo Yuan, Quanquan Gu, and Ilija Bogunovic.
\newblock Game-theoretic regularized self-play alignment of large language models, 2025.

\bibitem{geminiteam2024gemini15unlockingmultimodal}
Gemini Team, Petko Georgiev, Ving~Ian Lei, Ryan Burnell, et~al.
\newblock Gemini 1.5: Unlocking multimodal understanding across millions of tokens of context, 2024.

\bibitem{gemmateam2025gemma3technicalreport}
Gemma Team, Aishwarya Kamath, Johan Ferret, Shreya Pathak, et~al.
\newblock Gemma 3 technical report, 2025.

\bibitem{kimiteam2025kimivltechnicalreport}
Kimi Team, Angang Du, Bohong Yin, Bowei Xing, et~al.
\newblock Kimi-vl technical report, 2025.

\bibitem{meta2024llama32}
Meta Team.
\newblock Model cards \& prompt formats-llama 3.2, 2024.

\bibitem{opengv2024intern20}
OpenGVLab Team.
\newblock Vila: On pre-training for visual language models, 2024.

\bibitem{openai2024gpt4ocard}
OpenAI Team, Aaron Hurst, Adam Lerer, Adam~P. Goucher, et~al.
\newblock Gpt-4o system card, 2024.

\bibitem{qwen3}
Qwen Team.
\newblock Qwen3: Think deeper, act faster, April 2025.

\bibitem{vl-rethinker}
Haozhe Wang, Chao Qu, Zuming Huang, Wei Chu, Fangzhen Lin, and Wenhu Chen.
\newblock Vl-rethinker: Incentivizing self-reflection of vision-language models with reinforcement learning.
\newblock {\em arXiv preprint arXiv:2504.08837}, 2025.

\bibitem{wang2024piecingtogetherverifyingmultihop}
Haoran Wang, Aman Rangapur, Xiongxiao Xu, Yueqing Liang, Haroon Gharwi, Carl Yang, and Kai Shu.
\newblock Piecing it all together: Verifying multi-hop multimodal claims, 2024.

\bibitem{wang2024lvbench}
Weihan Wang, Zehai He, Wenyi Hong, Yean Cheng, Xiaohan Zhang, Ji Qi, Shiyu Huang, Bin Xu, Yuxiao Dong, Ming Ding, and Jie Tang.
\newblock Lvbench: An extreme long video understanding benchmark, 2024.

\bibitem{wang2025sotalessmctsguidedsample}
Xiyao Wang, Zhengyuan Yang, Chao Feng, Hongjin Lu, Linjie Li, Chung-Ching Lin, Kevin Lin, Furong Huang, and Lijuan Wang.
\newblock Sota with less: Mcts-guided sample selection for data-efficient visual reasoning self-improvement, 2025.

\bibitem{wang2025internvideo25empoweringvideomllms}
Yi Wang, Xinhao Li, Ziang Yan, Yinan He, Jiashuo Yu, Xiangyu Zeng, Chenting Wang, Changlian Ma, Haian Huang, Jianfei Gao, Min Dou, Kai Chen, Wenhai Wang, Yu Qiao, Yali Wang, and Limin Wang.
\newblock Internvideo2.5: Empowering video mllms with long and rich context modeling, 2025.

\bibitem{wang2025unifiedmultimodalchainofthoughtreward}
Yibin Wang, Zhimin Li, Yuhang Zang, Chunyu Wang, Qinglin Lu, Cheng Jin, and Jiaqi Wang.
\newblock Unified multimodal chain-of-thought reward model through reinforcement fine-tuning, 2025.

\bibitem{wei2023chainofthoughtpromptingelicitsreasoning}
Jason Wei, Xuezhi Wang, Dale Schuurmans, Maarten Bosma, Brian Ichter, Fei Xia, Ed Chi, Quoc Le, and Denny Zhou.
\newblock Chain-of-thought prompting elicits reasoning in large language models, 2023.

\bibitem{wei2025videoropemakesgoodvideo}
Xilin Wei, Xiaoran Liu, Yuhang Zang, Xiaoyi Dong, Pan Zhang, Yuhang Cao, Jian Tong, Haodong Duan, Qipeng Guo, Jiaqi Wang, Xipeng Qiu, and Dahua Lin.
\newblock Videorope: What makes for good video rotary position embedding?, 2025.

\bibitem{wu2024longvideobench}
Haoning Wu, Dongxu Li, Bei Chen, and Junnan Li.
\newblock Longvideobench: A benchmark for long-context interleaved video-language understanding, 2024.

\bibitem{wu2025stthinkmultimodallargelanguage}
Peiran Wu, Yunze Liu, Miao Liu, and Junxiao Shen.
\newblock St-think: How multimodal large language models reason about 4d worlds from ego-centric videos, 2025.

\bibitem{wu2024selfplaypreferenceoptimizationlanguage}
Yue Wu, Zhiqing Sun, Huizhuo Yuan, Kaixuan Ji, Yiming Yang, and Quanquan Gu.
\newblock Self-play preference optimization for language model alignment, 2024.

\bibitem{wu2024deepseekvl2mixtureofexpertsvisionlanguagemodels}
Zhiyu Wu, Xiaokang Chen, Zizheng Pan, Xingchao Liu, Wen Liu, Damai Dai, Huazuo Gao, Yiyang Ma, Chengyue Wu, Bingxuan Wang, Zhenda Xie, Yu Wu, Kai Hu, Jiawei Wang, Yaofeng Sun, Yukun Li, Yishi Piao, Kang Guan, Aixin Liu, Xin Xie, Yuxiang You, Kai Dong, Xingkai Yu, Haowei Zhang, Liang Zhao, Yisong Wang, and Chong Ruan.
\newblock Deepseek-vl2: Mixture-of-experts vision-language models for advanced multimodal understanding, 2024.

\bibitem{xiang2024atomthinkslowthinkingframework}
Kun Xiang, Zhili Liu, Zihao Jiang, Yunshuang Nie, Runhui Huang, Haoxiang Fan, Hanhui Li, Weiran Huang, Yihan Zeng, Jianhua Han, Lanqing Hong, Hang Xu, and Xiaodan Liang.
\newblock Atomthink: A slow thinking framework for multimodal mathematical reasoning, 2024.

\bibitem{xing2025echoinkr1exploringaudiovisualreasoning}
Zhenghao Xing, Xiaowei Hu, Chi-Wing Fu, Wenhai Wang, Jifeng Dai, and Pheng-Ann Heng.
\newblock Echoink-r1: Exploring audio-visual reasoning in multimodal llms via reinforcement learning, 2025.

\bibitem{xu2025redstardoesscalinglongcot}
Haotian Xu, Xing Wu, Weinong Wang, Zhongzhi Li, Da Zheng, Boyuan Chen, Yi Hu, Shijia Kang, Jiaming Ji, Yingying Zhang, Zhijiang Guo, Yaodong Yang, Muhan Zhang, and Debing Zhang.
\newblock Redstar: Does scaling long-cot data unlock better slow-reasoning systems?, 2025.

\bibitem{yang2024think}
Jihan Yang, Shusheng Yang, Anjali Gupta, Rilyn Han, Li Fei-Fei, and Saining Xie.
\newblock {Thinking in Space: How Multimodal Large Language Models See, Remember and Recall Spaces}.
\newblock {\em arXiv preprint arXiv:2412.14171}, 2024.

\bibitem{yang2025r1onevisionadvancinggeneralizedmultimodal}
Yi Yang, Xiaoxuan He, Hongkun Pan, Xiyan Jiang, Yan Deng, Xingtao Yang, Haoyu Lu, Dacheng Yin, Fengyun Rao, Minfeng Zhu, Bo Zhang, and Wei Chen.
\newblock R1-onevision: Advancing generalized multimodal reasoning through cross-modal formalization, 2025.

\bibitem{ye2024mplugowl3longimagesequenceunderstanding}
Jiabo Ye, Haiyang Xu, Haowei Liu, Anwen Hu, Ming Yan, Qi Qian, Ji Zhang, Fei Huang, and Jingren Zhou.
\newblock mplug-owl3: Towards long image-sequence understanding in multi-modal large language models, 2024.

\bibitem{yu2025dapoopensourcellmreinforcement}
Qiying Yu, Zheng Zhang, Ruofei Zhu, Yufeng Yuan, Xiaochen Zuo, Yu Yue, Tiantian Fan, Gaohong Liu, Lingjun Liu, Xin Liu, Haibin Lin, Zhiqi Lin, Bole Ma, Guangming Sheng, Yuxuan Tong, Chi Zhang, Mofan Zhang, Wang Zhang, Hang Zhu, Jinhua Zhu, Jiaze Chen, Jiangjie Chen, Chengyi Wang, Hongli Yu, Weinan Dai, Yuxuan Song, Xiangpeng Wei, Hao Zhou, Jingjing Liu, Wei-Ying Ma, Ya-Qin Zhang, Lin Yan, Mu Qiao, Yonghui Wu, and Mingxuan Wang.
\newblock Dapo: An open-source llm reinforcement learning system at scale, 2025.

\bibitem{zelikman2022starbootstrappingreasoningreasoning}
Eric Zelikman, Yuhuai Wu, Jesse Mu, and Noah~D. Goodman.
\newblock Star: Bootstrapping reasoning with reasoning, 2022.

\bibitem{zhang2025videollama3frontiermultimodal}
Boqiang Zhang, Kehan Li, Zesen Cheng, Zhiqiang Hu, Yuqian Yuan, Guanzheng Chen, Sicong Leng, Yuming Jiang, Hang Zhang, Xin Li, Peng Jin, Wenqi Zhang, Fan Wang, Lidong Bing, and Deli Zhao.
\newblock Videollama 3: Frontier multimodal foundation models for image and video understanding, 2025.

\bibitem{zhang2023videollamainstructiontunedaudiovisuallanguage}
Hang Zhang, Xin Li, and Lidong Bing.
\newblock Video-llama: An instruction-tuned audio-visual language model for video understanding, 2023.

\bibitem{zhang2025from}
Jiahui Zhang, Yurui Chen, Yanpeng Zhou, Yueming Xu, Ze Huang, Jilin Mei, Junhui Chen, Yujie Yuan, Xinyue Cai, Guowei Huang, Xingyue Quan, Hang Xu, and Li Zhang.
\newblock From flatland to space: Teaching vision-language models to perceive and reason in 3d.
\newblock {\em arXiv preprint arXiv:2503.22976}, 2025.

\bibitem{zhang2024longcontexttransferlanguage}
Peiyuan Zhang, Kaichen Zhang, Bo Li, Guangtao Zeng, Jingkang Yang, Yuanhan Zhang, Ziyue Wang, Haoran Tan, Chunyuan Li, and Ziwei Liu.
\newblock Long context transfer from language to vision, 2024.

\bibitem{zhang2025tinyllavavideor1smallerlmmsvideo}
Xingjian Zhang, Siwei Wen, Wenjun Wu, and Lei Huang.
\newblock Tinyllava-video-r1: Towards smaller lmms for video reasoning, 2025.

\bibitem{zhang2024videoinstructiontuningsynthetic}
Yuanhan Zhang, Jinming Wu, Wei Li, Bo Li, Zejun Ma, Ziwei Liu, and Chunyuan Li.
\newblock Video instruction tuning with synthetic data, 2024.

\bibitem{zhang2024openrftadaptingreasoningfoundation}
Yuxiang Zhang, Yuqi Yang, Jiangming Shu, Yuhang Wang, Jinlin Xiao, and Jitao Sang.
\newblock Openrft: Adapting reasoning foundation model for domain-specific tasks with reinforcement fine-tuning, 2024.

\bibitem{zhang2024mme}
Yi-Fan Zhang, Huanyu Zhang, Haochen Tian, Chaoyou Fu, Shuangqing Zhang, Junfei Wu, Feng Li, Kun Wang, Qingsong Wen, Zhang Zhang, et~al.
\newblock Mme-realworld: Could your multimodal llm challenge high-resolution real-world scenarios that are difficult for humans?
\newblock {\em arXiv preprint arXiv:2408.13257}, 2024.

\bibitem{zhang2024multimodalchainofthoughtreasoninglanguage}
Zhuosheng Zhang, Aston Zhang, Mu Li, Hai Zhao, George Karypis, and Alex Smola.
\newblock Multimodal chain-of-thought reasoning in language models, 2024.

\bibitem{zhao2025r1omniexplainableomnimultimodalemotion}
Jiaxing Zhao, Xihan Wei, and Liefeng Bo.
\newblock R1-omni: Explainable omni-multimodal emotion recognition with reinforcement learning, 2025.

\bibitem{zhao2025mmvu}
Yilun Zhao, Lujing Xie, Haowei Zhang, Guo Gan, Yitao Long, Zhiyuan Hu, Tongyan Hu, Weiyuan Chen, Chuhan Li, Junyang Song, Zhijian Xu, Chengye Wang, Weifeng Pan, Ziyao Shangguan, Xiangru Tang, Zhenwen Liang, Yixin Liu, Chen Zhao, and Arman Cohan.
\newblock Mmvu: Measuring expert-level multi-discipline video understanding, 2025.

\bibitem{zheng2025villavideoreasoningsegmentation}
Rongkun Zheng, Lu Qi, Xi Chen, Yi Wang, Kun Wang, Yu Qiao, and Hengshuang Zhao.
\newblock Villa: Video reasoning segmentation with large language model, 2025.

\bibitem{MLVU}
Junjie Zhou, Yan Shu, Bo Zhao, Boya Wu, Shitao Xiao, Xi Yang, Yongping Xiong, Bo Zhang, Tiejun Huang, and Zheng Liu.
\newblock Mlvu: A comprehensive benchmark for multi-task long video understanding.
\newblock {\em arXiv preprint arXiv:2406.04264}, 2024.

\end{thebibliography}

% %%%%%%%%%%%%%%%%%%%%%%%%%%%%%%%%%%%%%%%%%%%%%%%%%%%%%%%%%%%%

\appendix
\section{Detailed Training and Evaluation Analysis}

\subsection{Comparison of Training Speed}
\label{training_speed}
In this section, we compare the training time of the GRPO and DPO algorithms, both based on a single epoch of GRPO training.

For the first epoch of GRPO, the total dataset consists of approximately 47K samples. After discarding 80\% of the simpler examples, the dataset for the second epoch is reduced to around 24,653 samples. In contrast, the training data for DPO, after incorporating the \textit{Rollout-Aware Verifier}, comprises approximately 20,096 samples. The training process for both algorithms is conducted on 8 A800-80G GPUs, with the corresponding training time summarized in Table~\ref{tab:time_comparison}. The table reports the total training time in minutes, alongside the estimated average training time per sample, which is calculated by dividing the total training time by the number of samples. The average training time is presented in seconds.

From the results in Table~\ref{tab:time_comparison}, we observe that the average training time per sample for the GRPO algorithm is approximately 7 times longer than that of the DPO algorithm.
\begin{table}[ht]
\renewcommand{\arraystretch}{0.5}
  \caption{Training Time Comparison between DPO and GRPO.}
  \label{tab:time_comparison}
  \centering
  {%
  \begin{tabular}{lcc}
    \toprule
    % \textbf{Stage} & \textbf{\makecell{Reasoning \\ Activation}} & \textbf{\makecell{Video \\ Generalization}} & \textbf{DPO} & \textbf{GRPO2} \\
    \textbf{Stage} & \textbf{GRPO} & \textbf{DPO} \\
    \midrule
    \textbf{Framework} & OpenRLHF & trl \\
    \midrule
    \textbf{Size of Training Dataset} & 24,653 & 20,096\\
    \midrule
    \textbf{Total Training Time (minutes)} &  1891 &  242 \\
    \midrule
    \textbf{Sample-Level Training Time (seconds)} & 4.6 & 0.7 \\
    % \textbf{Gloabl Batch Size} & 128 & 64 & 32 & 64 \\
    % \textbf{Rollout Batch Size} & 64 & 64 & - & 64 \\
    % \textbf{Learning Rate} & 1e-6 & 1e-6 & 5e-7 & 5e-7 \\
    % \textbf{Rollout Responses per Query} & 8 & 8 & - & 8 \\
    % \textbf{Sampling Temperature} & 1.0 & 1.0 & - & 1.0 \\
    % \textbf{DPO Beta ($\beta$)} & - & - & 0.1 & - \\
    \bottomrule
  \end{tabular}
  }
\vspace{-12pt}
\end{table}

\subsection{Inconsistency Rate}
\label{inconsistency_rate}
The inconsistency rates for different models across the MMVU (mc), TOMATO, and Video-MMMU benchmarks are provided in Table~\ref{tab:inconsistency}.

\begin{table}[ht]
  \caption{Inconsistency Rate in Evaluation Benchmark}
  \label{tab:inconsistency}
  \centering
  \renewcommand{\arraystretch}{1.0}{%
  \begin{tabular}{lccc}
    \toprule
    \textbf{Model} & \textbf{MMVU (mc)} $\downarrow$ & \textbf{TOMATO} $\downarrow$ & \textbf{Video-MMMU} $\downarrow$ \\
    \midrule
    Qwen2.5-VL-7B (GRPO w cold start) & 6.4 & 11.9 & 9.7  \\
    \midrule
    VerIPO (Reasoning Activation) & 5.9 & 15.3 & - \\
    VerIPO (GRPO-Iteration1) & 13.3 & 17.4 & 15.7 \\
    VerIPO (DPO-Iteration1) & 8.0 & 11.0 & 13.4 \\
    VerIPO (GRPO-Iteration2) & 11.0 & 12.7 & 11.6 \\
    VerIPO (DPO-Iteration2) & 2.2 & 2.5  & 4.6\\
    VerIPO (DPO-Iteration2 no Consistency) & 6.7 & 7.6 & -\\
    VerIPO (DPO-Iteration2 no Reflection) & 5.3 & 7.0 & -\\
    VerIPO (GRPO-Iteration3) & 3.0  & 2.7 & 5.8 \\
    VerIPO (DPO-Iteration3) & 2.1 & 2.5 & 5.0\\
    \bottomrule
  \end{tabular} 
  }
  \vspace{-12pt}
\end{table}

\subsection{CoT Prompt}
\label{cot_prompt}
We have designed our prompt template based on the format used in DeepSeek-R1, where the system prompt explicitly defines the required output structure. This includes the use of <answer> tags to separate the reasoning process from the final answer. Detailed prompt are presented in Table~\ref{tab:prompt_setting}. The table lists two distinct prompt formats: one for multiple-choice questions and the other for numerical questions, where \{question\} represents the processed question.

\begin{table}[ht]
\renewcommand{\arraystretch}{0.5}
  \caption{Prompt setting for training and evaluation}
  \label{tab:prompt_setting}
  \centering
  {%
  \begin{tabular}{p{12cm}}
    \toprule
    Prompt For Multi-Choices Question \\
    \midrule
    \textbf{SYSTEM:} You should first thinks about the reasoning process in the mind and then provides the user with the answer. Your answer must be in latex format and wrapped in \$...\$.The reasoning process and answer are enclosed within <think> </think> and <answer> </answer> tags, respectively, i.e., <think> Since ...., so the answer is B. </think><answer> \$B\$ </answer>, which means your output should start with <think> and end with </answer>.\\
    \textbf{USER:} Question: \{question\}\\
    \midrule
    Prompt For Numberic Question \\
    \midrule
    \textbf{SYSTEM:} You should first thinks about the reasoning process in the mind and then provides the user with the answer. Your answer must be in latex format and wrapped in \$...\$.The reasoning process and answer are enclosed within <think> </think> and <answer> </answer> tags, respectively, i.e., <think> Since ...., so the answer is $2$. </think><answer> \$2\$ </answer>, which means your output should start with <think> and end with </answer>. \\
    \textbf{USER:} Question: \{question\} You must provide the answer in the <answer> </answer> tag, and the answer must be a number. \\
    \bottomrule
  \end{tabular}
  }
  % \vspace{-16pt}
\vspace{-12pt}
\end{table}

\subsection{Detailed Training Setting}
\label{datailed_training_setting}
During the training of Qwen2.5-VL-Instruct using the GRPO and DPO algorithms, we kept the visual encoder frozen throughout, training only the parameters of the MLP and the language model. For the GRPO training process, we utilized the Hybrid Engine to accelerate training. In the Reasoning Activation phase, both the \textit{micro train batch size} and \textit{micro rollout batch size} were set to 2. In the Group-Slow-Search phase, these values were reduced to 1 to accommodate the long video context inputs.

\subsection{Detailed Evaluation Setting}
\label{detailed_evaluation_setting}
When evaluating the Qwen2.5-VL-Instruct model, along with all models trained using reinforcement learning based on this architecture, we set do\_sample to False and used the default parameter settings from the Qwen generation\_config: repetition\_penalty = 1.05, temperature = 1e-6, and top\_p = 1.0. The entire evaluation process is accelerated by leveraging VLLM for inference.

For video sampling, we set the frame rate to 2.0 fps, configured the maximum number of sampled frames per video to 128, and specified the maximum resolution per frame as 256×28×28. Both the maximum number of sampled frames and the maximum resolution per frame were set to twice the values used during training. Additionally, we conducted a comparative experiment on the MMVU (mc) dataset and 300 long video samples sourced from Video-MME using the Qwen2.5-VL-Instruct model, with a focus on the number of sampled frames and the maximum resolution. The results of this experiment are presented in Table~\ref{tab:frames_experiment}.

\begin{table}[ht]
  \caption{Experiment about sampled frames and maximum resolution}
  \label{tab:frames_experiment}
  \centering
  \renewcommand{\arraystretch}{1.0}{%
  \begin{tabular}{lccccccccc}
    \toprule
    Model & FPS & Frames & Resolution & MMVU (mc) & Video-MME (Long-300)\\
    \midrule
    Qwen2.5-VL-7B (w.t.) & 1.0 & 64 & 128*28*28 & 57.9 & 54.0 \\
    Qwen2.5-VL-7B (w.t.) & 2.0 & 64 & 128*28*28 & 59.5 & 54.0 \\
    Qwen2.5-VL-7B (w.t.) & 2.0 & 64 & 256*28*28 & 61.0 & 49.7\\
    Qwen2.5-VL-7B (w.t.) & 2.0 & 128 & 128*28*28 & 61.0 & 51.3\\
    Qwen2.5-VL-7B (w.t.) & 2.0 & 128 & 256*28*28 & 63.0 & 53.0\\
    % Qwen2.5-VL-7B(w.t.) & 2.0 &\\
    \bottomrule
  \end{tabular} 
  }
  % \vspace{-16pt}
  % \vspace{-7pt}
\end{table}

% \subsection{}

\section{Qualitative Analysis}
\label{qualitativa_analysis}

\begin{figure}[ht]
    \centering
    \includegraphics[width=0.95\linewidth]{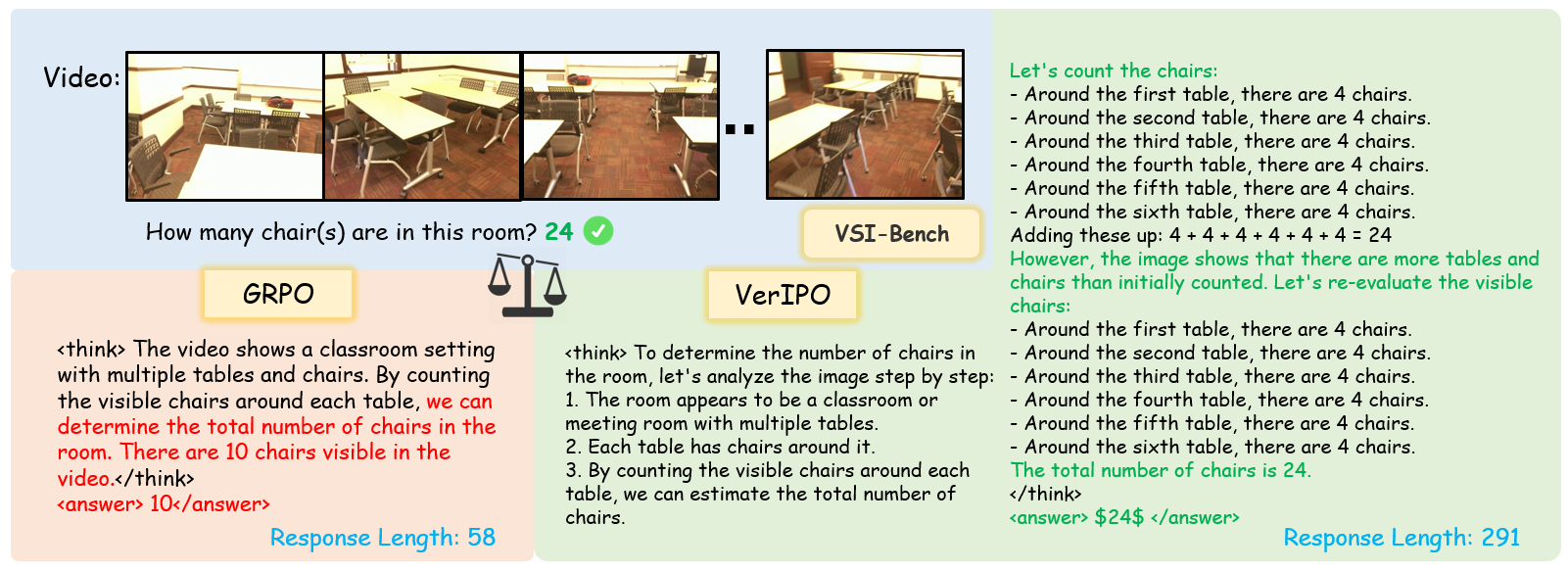}
    \caption{A case from VSI-Bench shows the comparative performance of GRPO and VerIPO. Our method is capable of generating longer responses and employing self-validation to address spatial reasoning tasks.}
    \label{fig:appendix_example1}
    % \vspace{-8pt}
\end{figure}

\begin{figure}[ht]
    \centering
    \includegraphics[width=0.95\linewidth]{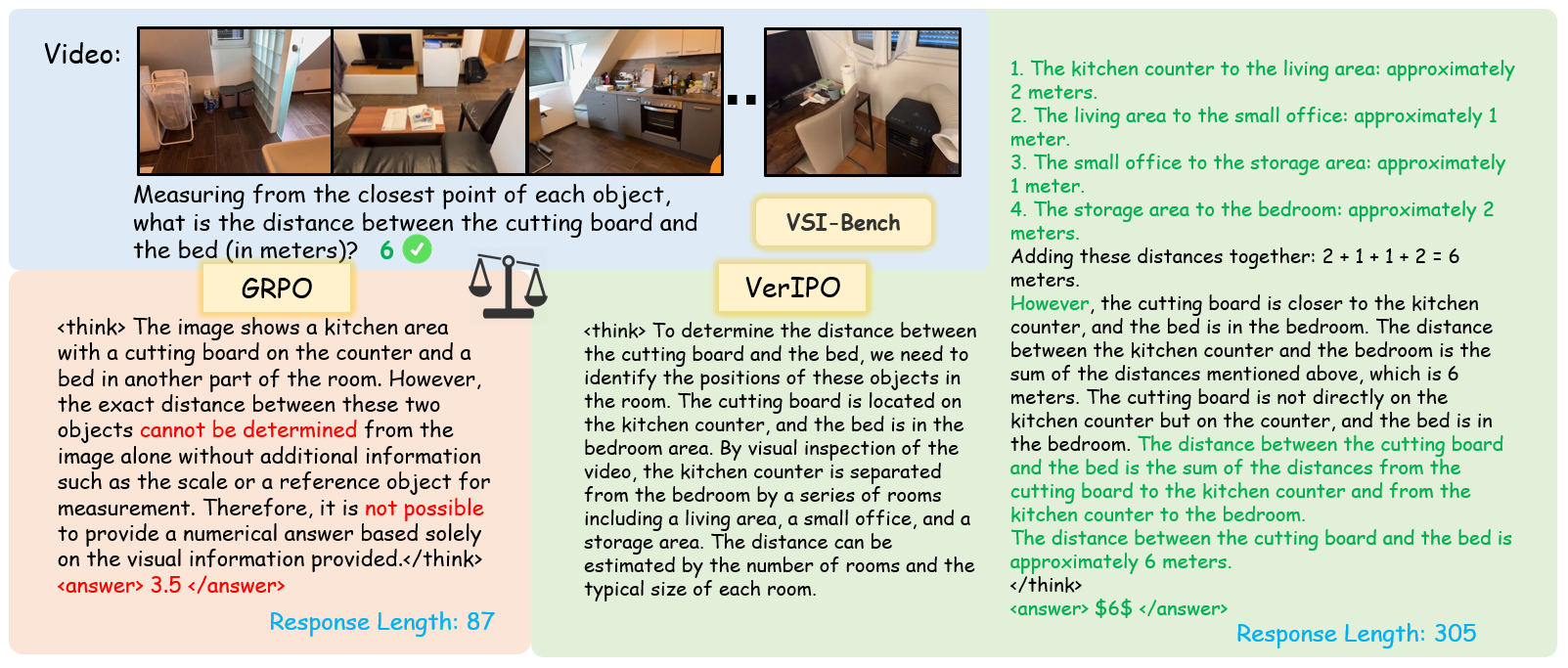}
    \caption{Another case from VSI-Bench shows the comparative performance of GRPO and VerIPO. Our method is capable of generating longer responses and employing self-validation to address spatial reasoning tasks.}
    \label{fig:appendix_example2}
    \vspace{-10pt}
\end{figure}

\begin{figure}[ht]
    \centering
    \includegraphics[width=0.95\linewidth]{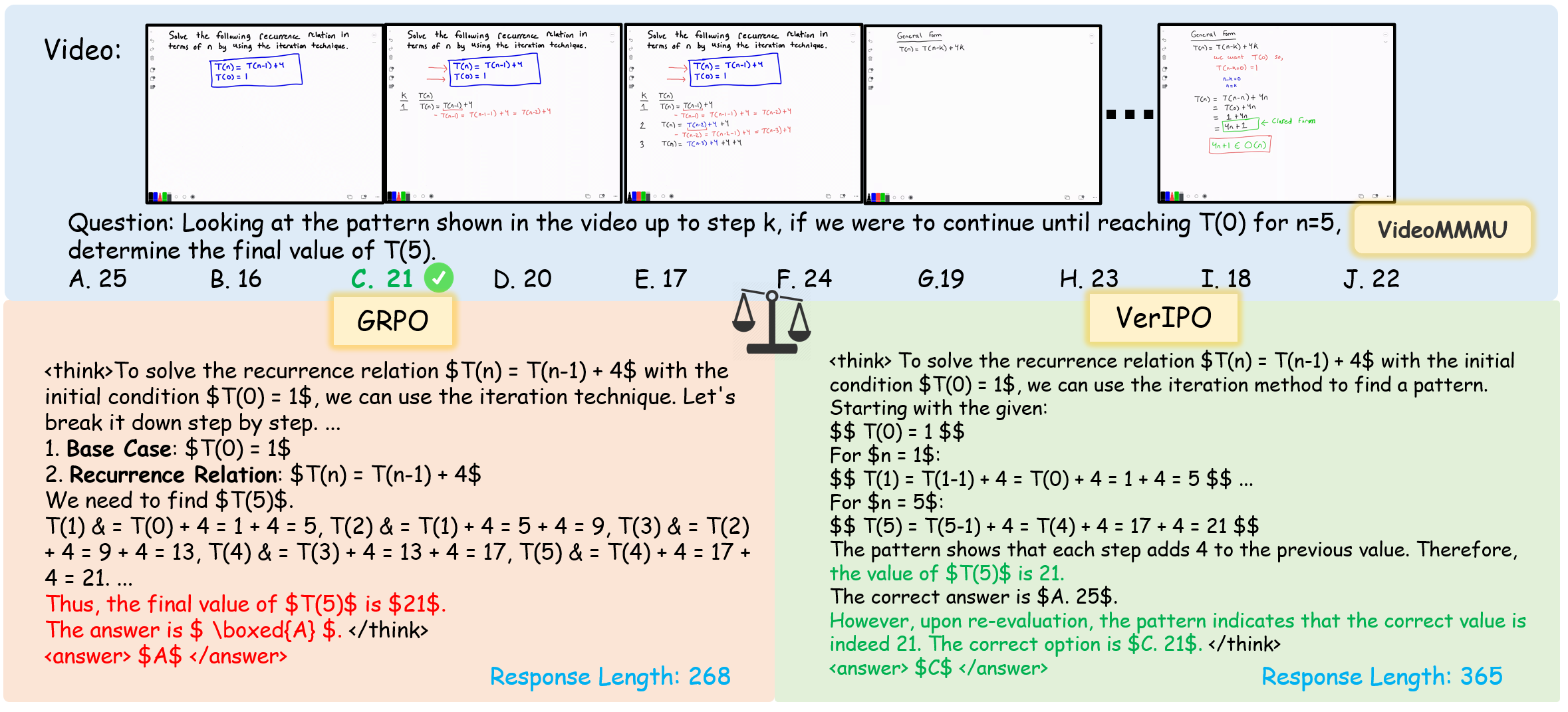}
    \caption{A case from Video-MMMU shows the comparative performance of GRPO and VerIPO. Our method can identify situations where the reasoning path is correct but an incorrect answer is chosen, through reflection, then re-selects the correct option that aligns with the reasoning content. }
    \label{fig:appendix_example3}
    \vspace{-10pt}
\end{figure}

\begin{figure}[!ht]
    \centering
    \includegraphics[width=1.0\linewidth]{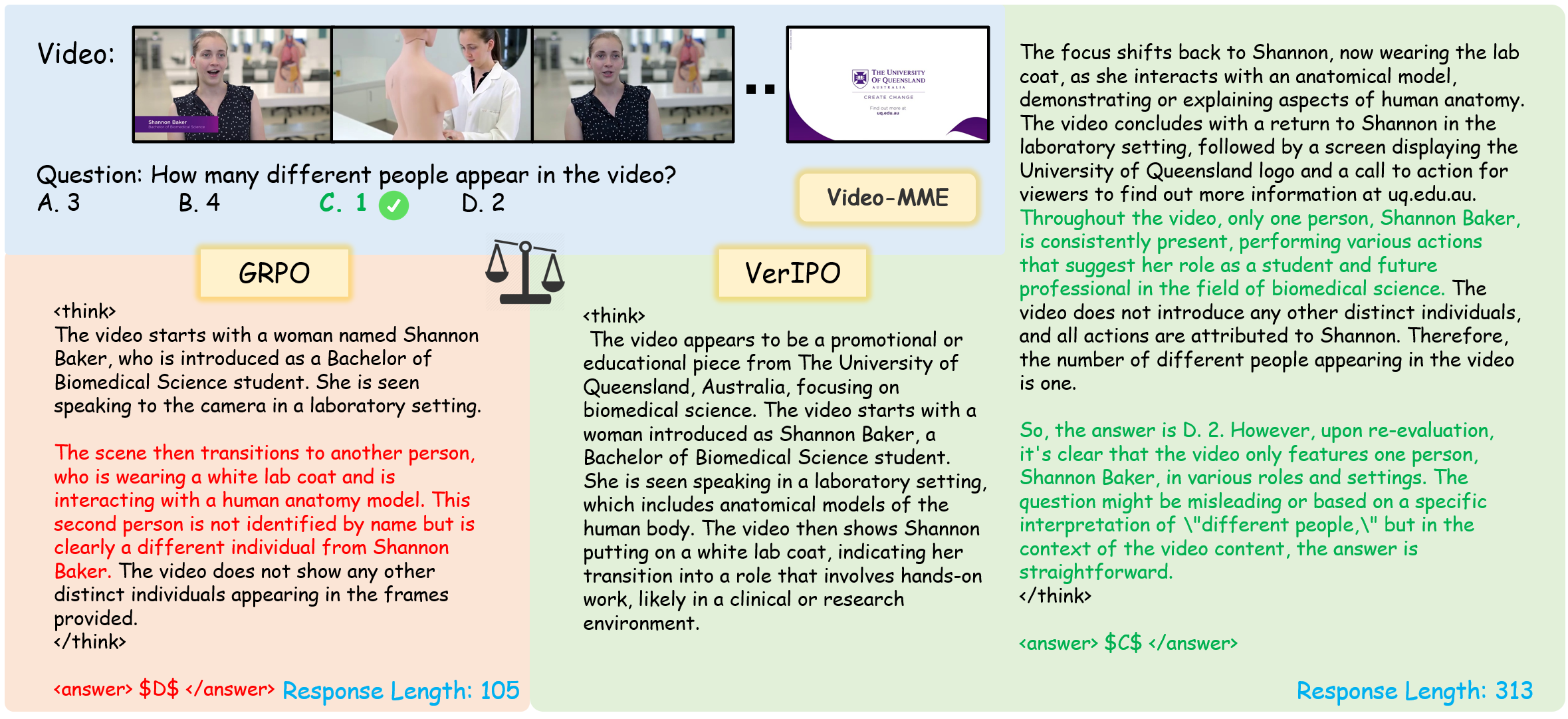}
    \caption{A case from Video-MME shows the comparative performance of GRPO and VerIPO. Our method also demonstrates strong capabilities in reflection and reasoning on general-domain question-answering tasks.}
    \label{fig:appendix_example5}
    \vspace{-10pt}
\end{figure}

\begin{figure}[ht]
    \centering
    \includegraphics[width=1.0\linewidth]{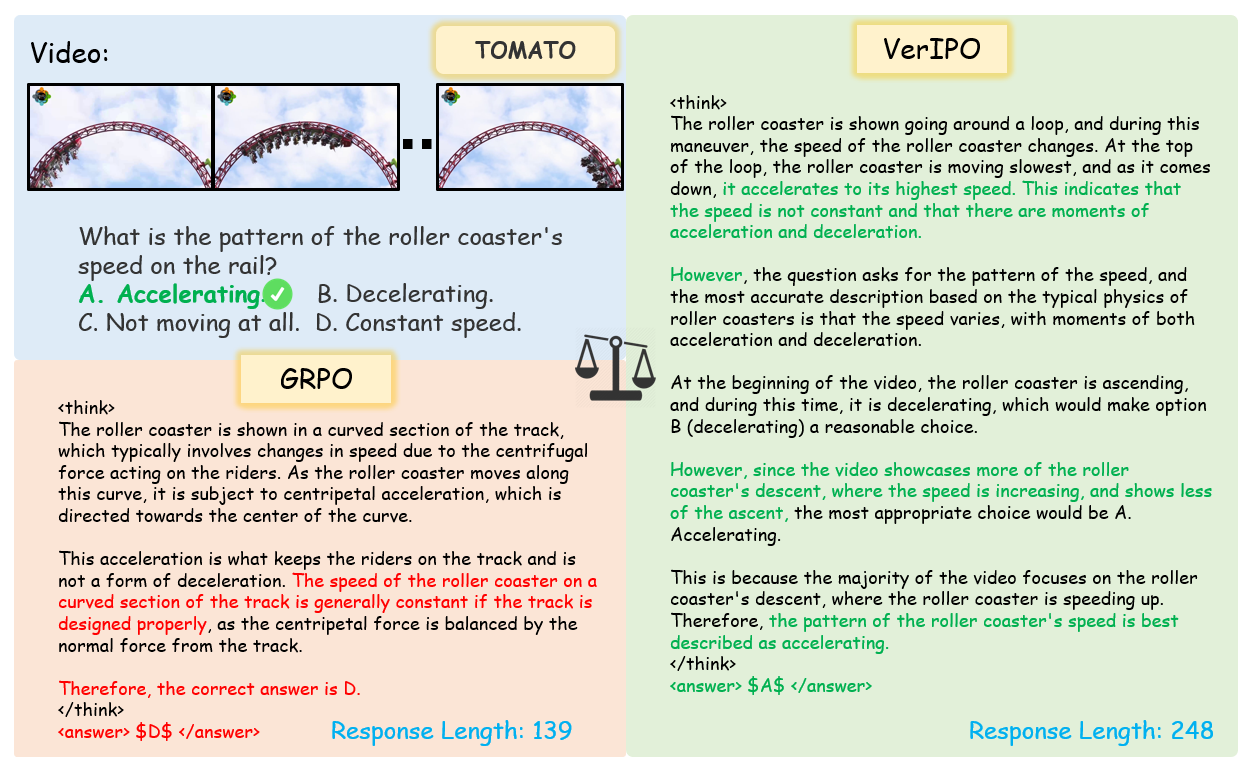}
    \caption{A case from TOMATO shows the comparative performance of GRPO and VerIPO. Our method is capable of generating longer responses and performing accurate temporal reasoning by self-validation.}
    \label{fig:appendix_example4}
    % \vspace{-8pt}
\end{figure}

% \section{Technical Appendices and Supplementary Material}
% Technical appendices with additional results, figures, graphs and proofs may be submitted with the paper submission before the full submission deadline (see above), or as a separate PDF in the ZIP file below before the supplementary material deadline. There is no page limit for the technical appendices.

% %%%%%%%%%%%%%%%%%%%%%%%%%%%%%%%%%%%%%%%%%%%%%%%%%%%%%%%%%%%%

%\newpage

\end{document}